\documentclass[10pt,journal,compsoc]{IEEEtran}
\ifCLASSOPTIONcompsoc
  \usepackage[nocompress]{cite}
\else
  \usepackage{cite}
\fi
\ifCLASSINFOpdf
\else
\fi
\usepackage[utf8]{inputenc}
\usepackage{amsmath}
\usepackage[most]{tcolorbox}
\usepackage{lipsum}
\usepackage{url}
 
\usepackage{graphicx} 
\usepackage{nccmath}
\usepackage{enumitem }
\usepackage{color,soul}
\usepackage{bm}
\renewcommand{\vec}[1]{\mathbf{#1}}
\usepackage{amsmath}
\usepackage[section]{placeins}
\graphicspath{ {images/} }
\usepackage{resizegather}
\usepackage[para,online,flushleft]{threeparttable}
\usepackage{setspace}
\usepackage{verbatim}
\usepackage{colortbl}

\usepackage{threeparttable}
\usepackage{multirow}

\newcounter{subeq}

\usepackage{subcaption}
\usepackage{hyperref}
\usepackage{tikz}
\usetikzlibrary{plotmarks}
\usepackage{pgfplots}
\usetikzlibrary{patterns}
\pgfplotsset{compat=1.12}
\everymath{\displaystyle}
\allowdisplaybreaks[1]
\makeatletter
\newcommand{\mathleft}{\@fleqntrue\@mathmargin\parindent}
\newcommand{\mathcenter}{\@fleqnfalse}
\makeatother

\usepackage{amssymb}
\usepackage[pass]{geometry}
\usepackage{algorithm}
\usepackage{diagbox}
\usepackage[export]{adjustbox}
\usepackage{algorithmic}
\usepackage{float}
\renewcommand\appendixname{}

\begin{document}
%
\title{Active Learning in Incomplete Label Multiple Instance Multiple Label Learning}
%
%
%
%

\author{Tam~Nguyen,~\IEEEmembership{Member,~IEEE,}
        and~Raviv~Raich,~\IEEEmembership{Senior Member,~IEEE}
\IEEEcompsocitemizethanks{\IEEEcompsocthanksitem T. Nguyen was with the School
of Electrical Engineering and Computer Science, Oregon State University, Corvallis,
OR, 97330-5501.\protect\\
E-mail: nguyeta4@oregonstate.edu
\IEEEcompsocthanksitem R. Raich  is with the School
of Electrical Engineering and Computer Science, Oregon State University, Corvallis,
OR, 97330-5501.\protect\\
E-mail: raich@eecs.oregonstate.edu}
\thanks{Manuscript received xxx; revised xxx.}}

%
%

\markboth{Tam Nguyen and Raviv Raich: ``Active Learning in Incomplete Label Multiple Instance Multiple Label Learning''}{Submitted to TKDE for review}

%



\IEEEtitleabstractindextext{%
\begin{abstract}

To alleviate labeling complexity, in multi-instance multi-label learning, each sample/bag consists of multiple instances and is associated with a set of bag-level labels leaving instances therein unlabeled. This setting is more convenient and natural for representing complicated objects with multiple semantic meanings. Compared to single-instance labeling, this approach allows for labeling larger datasets at an equivalent labeling cost. However, for sufficiently large datasets, labeling all bags may become prohibitively costly.  Active learning (AL) uses an iterative labeling and retraining approach to provide reasonable classification performance using a small number of labeled samples. To our knowledge, only two approaches have been previously proposed for AL in the MIML setting. These approaches either require labeling all classes in a selected bag or involve partial instance-level labeling. To further reduce labeling costs, we propose a novel bag-class pair-based approach for AL in the MIML setting. Due to the partial availability of bag-level labels, we focus on AL in the incomplete-label MIML setting. For the query process, we adapt AL criteria to the novel bag-class pair selection strategy. Additionally, we introduce an online approach for learning a discriminative graphical model based classifier. Numerical experiments on benchmark datasets demonstrate the effectiveness of the proposed approach.
\end{abstract}

\begin{IEEEkeywords}
Active learning, multiple instance multiple label learning, expected gradient length, uncertainty sampling, incomplete-label learning, bag-class pair.
\end{IEEEkeywords}}

\maketitle

\IEEEdisplaynontitleabstractindextext

%
\IEEEpeerreviewmaketitle

\IEEEraisesectionheading{\section{Introduction}\label{sec:introduction}}

%
%
%
%
\IEEEPARstart{I}{n} many real world applications, there are plentiful unlabeled
data but limited labeled data, and the acquisition of class labels is usually costly and difficult. By actively and iteratively selecting the most valuable data to query their supervised information, active learning tries to train an effective model with least labeling cost. Under the traditional single-instance single-label (SISL) learning, where each example is labeled by a single label, active learning
methods select the most valuable instances and then query their labels from the annotator (oracle) \cite{angluin1988queries, cohn1996active, king2004functional, king2009automation, cohn1994improving, dagan1995committee, mitchell1982generalization, seung1992query, dasgupta2008general, krishnamurthy2002algorithms, yu2005svm, fujii1998selective, thompson1999active, moskovitch2007improving, lewis1994sequential, mccallumzy1998employing, hoi2006large, tong2001support, zhang2002active, yang2003automatically, hauptmann2006extreme, tur2005combining, liu2004active}. The key task is to design a criterion for instance selection. In multi-instance learning (MIL), instances are grouped into bags which may contain any number of instances. A bag is labeled negative if and only if it contains all negative instances. A bag is labeled positive, however, if at least one of its instances is positive. There are two scenarios in active learning for MIL. The first is to simply allow the learner to query for the label of unlabeled bags \cite{salmani2014multi,zhang2010interactive}.  A second scenario is one in which all bags in the training set are labeled and the learner is allowed to query for the labels of instances selected from positive bags \cite{settles2008multiple}. In multi-label learning (MLL), where each example is labeled by a label set, there are several query approaches for active learning: (i) An instance is selected and the label for each class is obtained in a single query \cite{li2013active,tang2011semantic, li2004multilabel, hung2011multi}. (ii) An instance-class pair is selected to be labeled at each query \cite{wu2014multi, qi2008two}. (iii) An instance is selected first and then a class for the selected instance is labeled in each query \cite{huang2013active}. In MIML setting, every example is
represented by a bag of multiple instances and is annotated with multiple class labels to express the presence or absence of each class in the bag. The MIML setting provides an appropriate framework for learning with complex objects. However, when the instance feature vector dimension  and the number of classes increase, training an effective model requires more data. Moreover, since there are a large number of candidate labels in MIML, it becomes much more costly to annotated an example comparing to MIL. Hence, active learning for MIML is highly desired to reduce the labeling cost.
 \begin{figure*}[!t]
	\centering	\includegraphics[width=1.38\columnwidth]{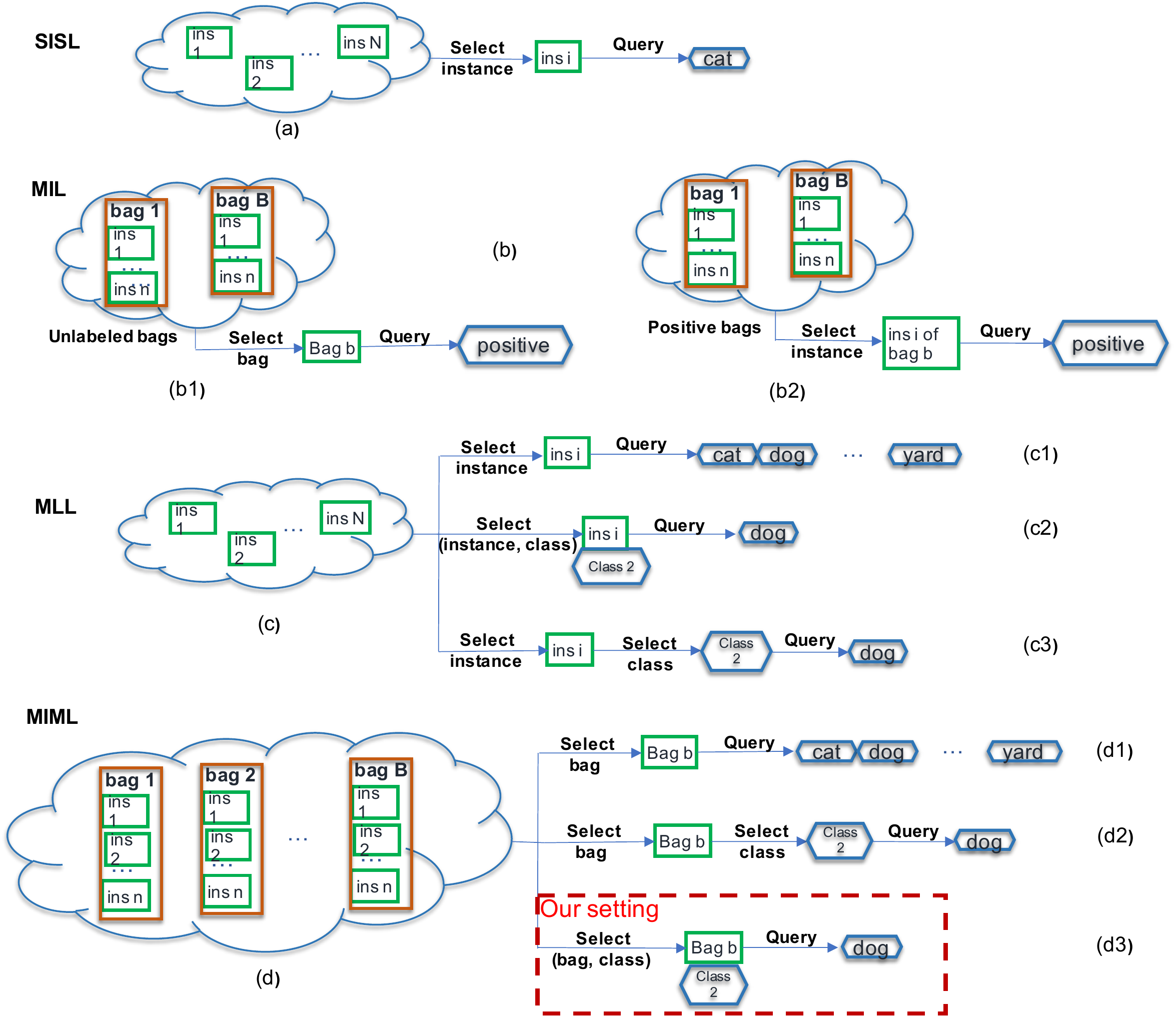}
	\caption{Active learning different settings. (a) Active learning in SISL setting. Each query selects the most informative instances from the unlabeled data to ask for labeling. (b) 
	Active learning in MIL setting. (b1) Each query selects a bag of instances to label. (b2) Each query selects an instance from a positive bag to label. (c) Active learning in MLL setting. (c1) Each query selects an instance to label all classes. (c2) Each query selects pair of instance and class to label. (c3) Each query selects one instance, and then select which class to label for selected instance. (d) Active learning in MIML setting. (d1) Each query selects a bag to label all classes. (d2) Each query selects one bag, and then select which class to label for selected bag.(d3) (\textbf{Our approach}) each query selects pair of bag and class to label. 
	}
	\label{fig:AL_SISL_MIML}
\end{figure*}
Similar to active learning in MLL, the query process for active learning in MIML learning presents several options. To the best of our knowledge, the only two approaches of querying in MIML, which are available in the literature are: (i) A bag is selected and all classes are labeled at each query \cite{retz2016active}. This approach may lead to redundant labeling of classes, which do not help to increase the performance of the model. (ii) A bag is selected first and then a single class for that bag is selected to be labeled at each query \cite{huang2017multi, wu2018predicting}. In \cite{huang2017multi, wu2018predicting}, after selecting the bag and the class to be labeled, the oracle decides whether they are relevant. If the oracle returns a negative label the query process is complete and all instances are assumed negative for the selected class. If the oracle returns a positive label, then the oracle is queried for an instance in the bag that is positive for the selected class. In either case, the information provided in each query can be directly mapped to instance labels. In turn, an instance level model is updated based on the specific instance information.
The queried bags are moved from the unlabeled set to the labeled set once all the labels of these bags are available. 
Beyond the standard cost for labeling a bag-class label pair, an additional cost is involved in the process of querying the labeler for the key instance associated with a positively-labeled class. To the best of our knowledge, there are no active learning methods for a single bag-class label pair query. This approach can be faster and/or less costly to label because in each query only one bag-class pair is presented to the labeler to label. The increase in labeling all classes in a given bag (e.g., the number of times the labeler reviews the bag-of-instances) compared to labeling a single class in the bag (i.e., a bag-class pair) may be small when the number of classes is small. If the number of classes considered is large, the cost of labeling  all classes in a bag compared to a single class may be significant, e.g., proportional to the number of classes considered. For example, in bird species recognition \cite{briggs2012acoustic}, a labeler can listen to an audio recording one time and determine the presence or absence of a single target species or even a small set of target species. However, if the number of target species is large, the labeler may need to repeatedly listen to the audio recording as they carefully review each class in their provided list of target species.
Though labeling a single bag-class pair can reduce the labeling cost, this approach introduces the following challenge. After a bag is labeled for one of its classes, the available label set for the bag is incomplete. Incomplete-label MIML is used to describe the setting, in which each training bags are provided with a subset of the correct label set.  Since up until recently no incomplete-label MIML methods were available, training based on data from the aforementioned active learning method was a challenge.  The recent work in \cite{nguyen2020incomplete} introduces a method to train data with sub-set of labels available. With the availability of training methods for incomplete MIML, we are now in position to evaluate the  bag-class label pair query paradigm for active learning in MIML. Fig.~\ref{fig:AL_SISL_MIML} presents active learning query types for different learning settings including SISL, MIL, MLL, and MIML. The active learning  strategy proposed in this paper is highlighted.\\
In summary, in this paper, we propose a novel framework for active learning under the MIML-ILL setting. Our MIML-ILL classifier bases on the discriminative graphical model with exact inference. We develop an online version of the model update to maximize the marginal log-likelihood to reduce the computational complexity of our framework and make it scalable. In the query process, we propose a novel approach to select bag-class pair based on EGL and uncertainty sampling that rely on the bag-class probability determined by our model. Finally, we build up a comprehensive comparisons  to show the effectiveness of the proposed approach.
\section{Related Work}
Active learning selectively queries the most valuable information from the oracle and aims to train an effective model with least queries. The key task in active learning is to design a proper strategy such that the queried information is most helpful for improving the learning model. There have been many active learning methods proposed under traditional single instance single label (SISL) setting \cite{settles2009active} 
with three main settings. (i) Membership query synthesis: the learner may request labels for any unlabeled instance in the input space, including (and typically assuming) queries that the learner generates de novo, rather than those sampled from some underlying natural distribution \cite{angluin1988queries,cohn1996active,king2004functional,king2009automation}. (ii) Stream-based selective sampling: the key assumption is that obtaining an unlabeled instance is free (or inexpensive), so it can first be sampled from the actual distribution, and then the
learner can decide whether or not to request its label \cite{cohn1994improving,dagan1995committee,mitchell1982generalization,seung1992query,dasgupta2008general,dagan1995committee,krishnamurthy2002algorithms,yu2005svm,fujii1998selective,thompson1999active,moskovitch2007improving}. (iii) Pool-based sampling: there is a small set of labeled data and a large pool of unlabeled data available. Queries are selectively drawn from the pool, which is
usually assumed to be closed \cite{lewis1994sequential,mccallumzy1998employing,hoi2006large,tong2001support,zhang2002active,yang2003automatically,hauptmann2006extreme,tur2005combining,liu2004active}.\\
While most active learning research focuses on traditional setting, there are a few works the extend the ideas to multi-instance learning \cite{settles2008multiple,zhang2010interactive,salmani2014multi} or multi-label learning \cite{li2004multilabel,hung2011multi,tang2011semantic,li2013active,wu2014multi,huang2013active}.  In \cite{settles2008multiple}, the authors introduce two active query selection strategies in multi-instance (MI) active learning: MI uncertainty sampling and
expected gradient length. They explore the case where a MI learner may query unlabeled instances from positively labeled bags in order reduce the inherent ambiguity of the MI representation. In \cite{salmani2014multi}, the authors describe a multiple-instance active learning algorithm for such incremental learning in the context of building models of relevant domain objects. Each bag of instance is selected to be labeled after each query. They introduce the concept of bag uncertainty sampling, enabling robots to identify the need for feedback, and to incrementally revise learned object models by associating visual cues extracted from images with verbal cues extracted from limited high-level human feedback.
Two general multiple-instance active learning (MIAL) methods are introduced in \cite{zhang2010interactive}, multiple-instance active learning with a simple margin strategy (S-MIAL) and multiple-instance active learning with Fisher information (F-MIAL). These two approaches are applied to the active learning in localized content based image retrieval. S-MIAL considers the most ambiguous picture as the most valuable one, while F-MIAL utilizes the Fisher information and analyzes the value of the unlabeled pictures by assigning different labels to them. Both methods select a bag to label for each query. For active learning in the MLL setting, 
the approaches in \cite{li2013active,tang2011semantic,li2004multilabel,hung2011multi} select a single instance to label all classes after each query. In \cite{li2013active}, the authors first propose two novel multi-label active learning strategies, a max-margin prediction uncertainty sampling strategy and a label cardinality inconsistency strategy, and then integrate them into an adaptive framework of multi-label active learning. In \cite{li2004multilabel}, the authors propose a multi-label SVM active learning method. They provide two selection strategies: max loss strategy and mean max loss strategy. Auxiliary learner is introduced in \cite{hung2011multi}. They extend maximum loss reduction with maximum confidence (MMC) to a more general framework that removes the heavy dependence and clarifies the roles of each component in MMC. In particular, the framework is characterized by a major learner for making predictions, an auxiliary learner for helping with query decisions and a query criterion based on the disagreement between the two learners. In \cite{tang2011semantic}, the authors propose a semantic-gap-oriented active learning method, which incorporates the semantic gap measure into the information-minimization-based sample selection strategy. The basic learning model used in the active learning frame-work is an extended multi-label version of the sparse-graph-based semi-supervised learning method that incorporates the semantic correlation. The different strategy of active learning in MLL is introduced in \cite{wu2014multi,qi2008two}. In these approaches, a pair of instance-label is selected simultaneously to label after each query. Specifically, in \cite{wu2014multi}, the authors propose a novel example-label based multi-label active learning method. They consider how to select the most informative example-label pairs by computing the uncertainty of each example-label pair with the boundary, but they did not take the label correlation of an example into consideration. In \cite{qi2008two}, the authors propose to select sample-label pairs to minimize a multi-label Bayesian classification error bound. This active learning strategy not only considers the sample dimension but also the label dimension and is termed Two-Dimensional Active Learning (2DAL). In \cite{huang2013active}, the authors propose an approach to select a single instance first and then a class for the selected instance is labeled after each query. In this approach, the selected instance is the one that maximizes the label cardinality inconsistency (LCI). LCI measures the inconsistency between the number of predicted positive labels of an instance and the average label cardinality (the average of the number of positive lablels) on the fully labeled data. And then a class is selected based on the distance between its and the dummy label.
All the above studies are focusing on either multi-instance or multi-label learning, and cannot be directly applied to MIML setting. There are several studies which are developed for active learning in MIML setting. The method in \cite{retz2016active} is specifically designed based on MIMLSVM. It firstly degenerates the bags to single-instance representation and then directly employ traditional active learning method for label querying (select each bag to label all classes for each query), which does not truly exploit the characteristics of MIML tasks. 
The authors in \cite{huang2017multi} propose an approach for active learning in MIML setting based on the work in \cite{huang2013active}.  For each query, the bag is selected first based on the uncertainty (the gap between the predicted number of positive labels of the bag and the average number of positive labels of the training data) and diversity of the bag (how many labels of the bag was queried before). Then a class is pointed out to be labeled for the selected bag based on the distance from the label to the thresholding
dummy label. In their methods, not only one selected label is queried, but also the key instance which is most relevant to queried label is asked. In \cite{wu2018predicting}, the authors extend work in \cite{huang2017multi} with a modification in bag label prediction achieving from instance-level predictions to predicting protein functions of bacteria genomes. All these approaches select a bag to label all classes or a bag and then select a class to label. To the best of our knowledge, there are no studies that focus on selecting a bag-class pair in the MIML setting. 
\section{The proposed approach}
In this section, we introduce the problem of active learning for MIML data with missing labels. We present a novel instance selection approach,  in which the presence or absence of a specific class in a given bag is obtained with each query. We demonstrate how criteria such as expected gradient length (EGL) and uncertainty sampling, commonly developed for querying a multi-class label for the  single instance case, can be modified for the selection of the bag-class pair as an instance. Finally, to facilitate an efficient model update, we develop an online SGD approach for learning the model parameters.
\subsection{Problem formulation}
Our main goal is to develop a model to learn an effective classifier that can label a newly unseen bag/instance under the setting of MIML learning with missing labels with as smallest as possible number of training data using active learning. We begin with a description of the data and related notation. We then continue with the probabilistic model for this setting and the associated inference approaches.\\
\noindent{\it Data description:}~~
 We consider an entire dataset consisting of a collection of $B$ bags and their associated label sets $\{(\vec{X}_b,\vec{Y}_{b}) \}_{b=1}^B $, respectively.
Each bag $\vec{X}_b$ is a set of instance feature vectors, $\vec{X}_b = \{ \vec{x}_{b1}, \vec{x}_{b2}$, \ldots , $\vec{x}_{bn_b} \}$, where $ \vec{x}_{bi} \in  {\cal X} \subseteq  R^d $  is the feature vector for the $ith$ instance in the $bth$ bag and $n_b$ denotes the number of instances in the $bth$ bag.  Bag $b$ is labeled by a label vector $\vec{Y}_b \in \{-1,0,1\}^C$, where $C$ is the number of classes and for each class the $cth-$entry $Y_{bc} \in \{-1,0,1 \} $ indicates a positive label $1 $, negative label $0$, and the  absence  of the label $-1$. The set of available labels in bag $b$ is denoted by $S_b$, which is defined as:
\begin{equation}\label{eq:Sb_definition}
S_b=\{c|Y_{bc} \neq -1,c = 1, 2, \dots, C\}.
\end{equation}
Additionally, we introduce the set of positively labeled classes  $S_b^+$, negatively labeled classes $S_b^-$, and unlabeled classes $\bar{S_b}$ in bag $b$:
\begin{equation}
\begin{aligned}
&S^+_b=\{c|Y_{bc} = 1,c = 1, 2, \dots, C\}\\
&S^-_b=\{c|Y_{bc} = 0 ,c = 1, 2, \dots, C\}.\\
&\bar{S}_b=\{c|Y_{bc} = -1,c = 1, 2, \dots, C\}.
\end{aligned}
\end{equation}
Note that $\textstyle{S_b = S_b^+ \cup S_b^-}$ and $\bar{S_b}$ is the complement of $S_b$. For example, let $C=6$ and $Y_b=[0,-1,1,1,0,-1]^T$. Hence, we only observe a label for $Y_{b1}$, $Y_{b3}$, $Y_{b4}$ and $Y_{b5}$. Therefore, $S_b=\{1, 3, 4, 5\}$, $S_b^+=\{3, 4\}$, $S_b^-=\{1, 5\}$, and $\bar{S_b}=\{2, 6\}$.\\
Moreover, we can define the set ${\cal L}$, the available label index set, as the set of indices $(b,c)$ for which  $Y_{bc} \neq -1$, i.e., ${\cal L}=\{(b,c)|Y_{bc} \neq -1\}$. Similarly, we can define the unavailable label index set, as the set of indices $(b,c)$ for which $Y_{bc} = -1$, ${\cal U}=\{(b,c)|Y_{bc} = -1 \}$. We consider two steps in active learning: (1) instance selection and (2) model update. An instance selection criterion is applied to obtain bag-class pair $(b^*,c^*) \in {\cal U}$ for which a label will be provided. After a label is provided for such an instance, the corresponding $Y_{b^*c^*}$ will no longer be $-1$ and consequently ${\cal L} \leftarrow {\cal L} \cup \{(b^*,c^*)\}$ and  
${\cal U} \leftarrow {\cal U} \backslash \{(b^*,c^*)\}$. For the model update, it is common to retrain the model after obtaining an additional label. It is important to note that using our notations the training set remains $\{(\vec{X}_b,\vec{Y}_b)\}$ regardless of how many unavailable labels. Their availability is directly encoded in the $\vec{Y}_b$ vector. With more bag-class pairs becoming known, some of the entries in $Y_{bc}$ change from $-1$ to either $1$ or $0$. Though bags for which $\vec{Y}_b$ is the all $-1$ vector are also included in the training set, they play no role in the update step since the log-likelihood will only be computed based on the available labels.\\
Given this setting, we develop a novel framework of active learning in MIML-ILL. In which, (i) we adopt the MIML-ILL model from \cite{nguyen2020incomplete} (as reviewed in this section) to learn the base classifier; (ii) we propose a novel instance selection, called bag-class pair selection based on EGL and uncertainty sampling for the MIML-ILL setting to select the most informative bag-class pair from the unlabeled data to update the model; (iii) we present an online version of optimization problem to maximize the marginal log-likelihood of the model in \cite{nguyen2020incomplete}. The complete algorithm of our proposed approach is summarized in   
Algorithm~\ref{alg:AL-alg}.
{\color{blue}\begin{algorithm} 
	\scriptsize
	\caption{The MIMLILL-AL algorithm}\label{alg:AL-alg}
	\begin{algorithmic}[1]
		\STATE \textbf{Input:}\\
		${\cal B} =\{(\vec{X}_b,\vec{Y}_{b}) \}_{b=1}^B $: entire dataset (may include incomplete label vectors).\\
		${\cal U} = \{(b,c)|Y_{bc}=-1\}$: the  unavailable bag-class pairs;\\
		$Q$: number of queries; (no more than $|{\cal U}|$)\\
		\STATE \textbf{Initialize:}\\
		$q=1$\\
		$\vec{w}_0=  0$\\
		$\vec{w}=  Model\_Update({\cal B},\vec{w}_0)$ (as in Section \ref{subsec:background})
		\WHILE{$q \leq Q$}  
		\STATE $(b^*,c^*) = Instance\_Selection({\cal B},{\cal U}, \vec{w})$ ~~~(as in Section \ref{sec:ins_selection})\\
		\STATE Query $Y_{b^*c^*}$\\
		\STATE Update training data:\\
        ${\cal U} \leftarrow {\cal U} \backslash \{(b^*,c^*)\}$.\\
		\STATE $\vec{w} = Model\_Update({\cal B},\vec{w})$~~~(as in Section \ref{sec:model_update})\\
		\STATE $q = q + 1$
		\ENDWHILE
	\end{algorithmic}
\end{algorithm}}
\subsection{Background}\label{subsec:background}
\begin{figure}[!]
	\centering	\includegraphics[width=.3\columnwidth]{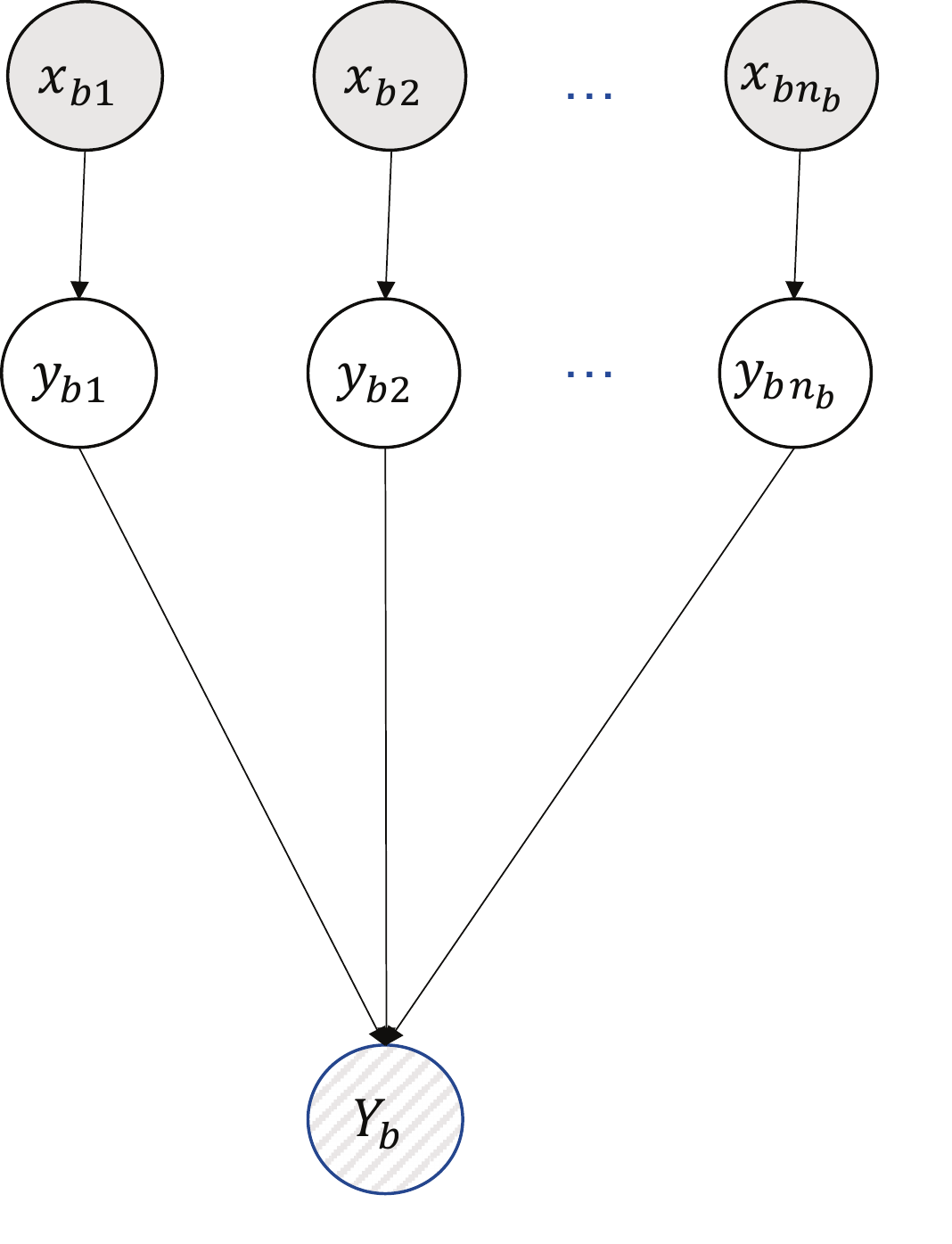}
	\caption{Incomplete label learning model. Shaded nodes are observed (i.e., $\vec{x}_{b1},\ldots,\vec{x}_{bn_b}$) and striped nodes are partially-observed (i.e., $\vec{Y}_b$). }
	\label{fig:original-reformMIML}
\end{figure}

Our proposed model is adopted from \cite{nguyen2020incomplete} paper (presented in Fig.~\ref{fig:original-reformMIML}). Specifically, we assume that the instance feature vectors and the bag labels in each bag are independent across bags, i.e., observations $(\vec{X}_b,\vec{Y}_b)$ are independent for $b=1,2,\ldots,B$.  To define the model for a single bag, we assume that the latent multi-class instance labels $y_{bi}$ for $i=1,2,\ldots,n_b$ are independent conditioned on ${\bf X}_b$ and the probability for $y_{bi} \in \{1,2,\ldots,C\}$ given $\vec{x}_{bi}$ follows the multinomial logistic regression model
\begin{equation}\label{eq:regression1}
P(y_{bi}=c|\vec{x}_{bi},\vec{w}) =\frac{e^{\vec{w}_c^T\vec{x}_{bi}}}{\sum_{k=1}^{C} e^{\vec{w}_k^T\vec{x}_{bi}}},
\end{equation}
where $\vec{w}=[\vec{w}_1^T,\ldots,\vec{w}_C^T]^T$ is the model parameter column vector and $\vec{w}_c$ for $c=1,2,\ldots,C$ is a $d$-dimensional column vector. As discussed in the introduction, due to various reasons (e.g., labeling cost and/or labeling strategy), only a subset of the classes $\{1, 2, \dots, C\}$ may be labeled. We model the labeled (i.e., observed) entries of $\vec{Y}_b$ conditioned on $\vec{y}_b = [y_{b1}, \dots, y_{bn_b}]$ as independent and consequently $\textstyle{ P(\vec{Y}_b|\vec{y}_b) = \prod_{c \in S_b}P(Y_{bc}|\vec{y}_b)}$. To model $Y_{bc}$ for $c \in S_b$ given $\vec{y}_b$, we follow the OR rule assumption, i.e., $Y_{bc}=1$ if at least one of the $y_{bi}$'s equals to $c$ and $0$ otherwise. Hence,
we define the probability of $Y_{bc}$ for $c \in S_b$ given $\vec{y}_b$ as:
\begin{equation}\label{pYc}
P(Y_{bc}=Y|\vec{y}_b) = Y (1 - \prod_{i=1}^{n_b} I(y_{bi} \neq c)) + (1-Y) \prod_{i=1}^{n_b} I(y_{bi} \neq c), 
\end{equation}
where $Y \in \{0,1\}$. Based on the aforementioned modeling assumptions, the single bag log-likelihood for the $b$th bag is given by
\begin{equation}\label{eq:ori_loglikelihood_prior}
\begin{aligned}
\log P({\bf Y}_b,{\bf X}_b) = \log \sum_{\vec{y}_b} \prod_{c\in S_b}P(Y_{bc}|\vec{y}_b)\prod_{j=1}^{n_b} P(y_{bj}|\vec{x}_{bj},\vec{w}) +\log P({\bf X}_b),
\end{aligned}
\end{equation}
where $P(y_{bj}|\vec{x}_{bj},\vec{w})$ and $P(Y_{bc}|\vec{y}_b)$ are given by (\ref{eq:regression1}) and  (\ref{pYc}), respectively. Note that term $\log P({\bf X}_b)$ is not a function of the parameter vector $\vec{w}$ and hence is treated as a constant. \\ 
According to \cite{nguyen2020incomplete}, the computational complexity associated with the E-step of EM algorithm used to maximize the log-likelihood function is the exponential degree of the number of positive labels per bag. To reduce the computational complexity associated with the E-step when the number of positive labels per bag is large, we adopted the marginal maximum likelihood (MML) approach to maximum likelihood. We consider an objective that is the sum of the log-likelihoods associated with each class label. The contribution of each class can be handled separately. Specifically, the use of a single label at a time allows us to compute the objective function in closed-form and implement its gradient efficiently, thereby reducing the complexity to linear in the number of classes at the potential expense of breaking down some of the label dependence relations.\\
Based on the model, the sum of marginal log-likelihoods is:
\begin{eqnarray}\label{eq:MMLobj}
	L_{MML}(\vec{w}) =\sum_{b=1}^B \sum_{t\in S_b}\log P(Y_{bt} | \vec{X}_b,\vec{w}).
\end{eqnarray}
We can further derive the criterion by marginalizing $P(Y_{bt},\vec{y}_b|\vec{X}_b,\vec{w})=P(Y_{bt}|\vec{y}_b)\textstyle{\prod_{i=1}^{n_b}} P(y_{bi}|\vec{x}_{bi},\vec{w})$ over $\vec{y}_b$ while substituting $P(Y_{bt}|\vec{y}_b)$ from (\ref{pYc}) \cite{nguyen2020incomplete}. The aforementioned step along with some simplifications yields the follow expression for the marginal log-likelihood objective:
\begin{equation}\label{eq:mml_final_function}
\begin{aligned}
	L_{MML}(\vec{w}) =\sum_{b=1}^B \sum_{t\in S_b}&\big[ I(Y_{bt}=0) \sum_{i=1}^{n_b} \log P(y_{bi} \neq t) + \\
	&I(Y_{bt}=1) \log  (1 - e^{\sum_{i=1}^{n_b} \log P(y_{bi} \neq t)})\big].
\end{aligned}
\end{equation}  
For $c=1,\ldots,C$, the update rule of $\vec{w}_c$ is given by
\begin{equation}\label{eq:gdupdate}
\begin{aligned}
\vec{w}_{c}^{k+1} = \vec{w}_{c}^k +\eta_k   \frac{\partial L_{MML}(\vec{w})}{\partial \vec{w}_{c}}|_{\vec{w}=\vec{w}^k}, 
\end{aligned}
\end{equation}
where $\textstyle{ \frac{\partial L_{MML}(\vec{w})}{\partial \vec{w}_{c}}}$ is computed as
\begin{equation}\label{eq:grad_MML}
\begin{aligned}
\sum_{b=1}^B\sum_{i=1}^{n_b} \sum_{t \in S_b}& \biggl(
I(Y_{bt}=1) P(Y_{bt}=0)\frac{P(y_{bi}=t)}{P(Y_{bt}=1)}- I(Y_{bt}=0)P(y_{bi}=t)  \biggr) \\
&  \biggl(I(c=t) - I(c\neq t)\frac{P(y_{bi} = c)}{1-P(y_{bi} = t)}\biggr) \vec{x}_{bi}
\end{aligned}
\end{equation}\vspace{-0.8cm}
and 
\begin{equation}\label{eq:bag_neg}
\begin{aligned}
    &P(Y_{bt} =0) = \prod_{i=1}^{n_b} p(y_{bi} \neq t),\\
    &P(Y_{bt} =1) = 1 - P(Y_{bt} =0).
\end{aligned}  
\end{equation}
The step size $\eta_k$ is determined using the backtracking line search algorithm. 

We develop our framework for active learning under the MIML-ILL setting based on this model. The aforementioned inference (\ref{eq:gdupdate})-(\ref{eq:bag_neg}) was proposed in \cite{nguyen2020incomplete} as means of training the model-based MIML-ILL classifier. In active learning, the model is typically updated with every query to ensure that instance selection is taking into account the most current and accurate model. This task is computationally intensive with a computational complexity that grows linearly with the the size of the labeled data. To reduce the computational complexity of our framework and make it scalable, we develop an online version of the model update to maximize the marginal log-likelihood (the detail is provided in Section \ref{sec:model_update}). In the query process, we propose a novel approach to select bag-class pair based on EGL and uncertainty sampling that rely on the bag-class probability determined by this model. Details are provided in Section \ref{sec:ins_selection}.
\subsection{Instance Selection}\label{sec:ins_selection}
Recall an unavailable label index set, as the set of indices $(b,c)$ for which $Y_{bc} = -1$, ${\cal U}=\{(b,c)|Y_{bc} = -1 \}$, where $Y_{bc} \in \{-1, 0, 1\}$ and $c = 1, \dots, C$. Let $l \in \{0, 1\}$. In our active learning scenario, each query selects a pair of bag-class $(b,c)$ in $\cal U$ and queries for $Y_{bc} = l$. We introduce two criterion to select bag-class pair from the unlabeled pool: (i) EGL and (ii) uncertainty sampling.\\
\subsubsection{EGL for MIML} 
Expected gradient length is an approach used for discriminative probabilistic model classes. This approach selects the instance that would impart the greatest change to the current model if we knew its label. Specifically, the learner should query instance $\vec{x}$ which, if labeled and added to training data ${\cal L}$, would result in the new training gradient of the largest magnitude. Let $f_{\theta}({\cal L})$ be the objective function used in training a classifier (e.g., log-likelihood or regularized loss). and $\nabla f_{\theta}({\cal L})$ be the gradient of the objective function $f$ w.r.t $\theta$. Now let $\nabla f_{\theta}({\cal L}\cup \{(\vec{x},y)\})$ be the new gradient obtained by adding the training tuple $(\vec{x},y)$ to the previously available data ${\cal L}$. Since the query algorithm does not know the true label $y$ in advance, we must instead calculate the length as an expectation over the possible labelings:
\begin{equation}
    \vec{x}^*_{EGL} = \arg \max_{\vec{x}\in {\cal U}} \sum_y P(y|\vec{x},\vec{w})\|\nabla f_{\theta}({\cal L}\cup \{(\vec{x},y)\})\|. 
\end{equation}
This EGL approach is applied in single-instance single-label (SISL) or multi instance learning (MIL). For multi-label learning (MLL), to consider all label sets of one instance, we need to sum of gradient of $2^C$ possible cases, where $C$ is the number of classes. This leads to issue of computational complexity when $C$ is large. However, since we restrict our attention to selecting the bag-class pair, we manage to avoid the high computation cost associated with the calculation of the EGL criterion for an entire bag. Moreover, with selecting the bag-class pair, we might also reduce the redundant labeled classes due to query the labels of the entire bag.\\
To that end, we develop a novel extension of the EGL criterion to the MIML-ILL setting to select a bag-class pair for labeling during the query phase. Given the model parameters and the current training data, for each bag-class pair in the unlabeled data, we compute the gradient of the marginal log-likelihood (\ref{eq:mml_final_function}) associated with the training data after a new bag-class pair is included. The bag-class pair that has the most impact to the gradient of the marginal log-likelihood on the training data with this pair included is selected for the label querying. This pair is then added to the current training data for model update. \\
Specifically, given the current label index set  ${  \cal L} = \{(b,c)|Y_{bc} \neq -1\}$ and the current model parameter vector $\vec{w}$, for each pair of bag-class $(b,c)$ in ${\cal U}$, we add this pair to current available index set ${\cal L}$, then compute the gradient of the marginal log-likelihood (ML) on this new available index set. The number of available indices in ${\cal L}$ will increase by $1$ and the gradient of the marginal log-likelihood for new available index set is:
\begin{equation}
    \begin{aligned}
    G_{{\cal L}\cup\{(b,c)\}} &= \frac{1}{|{\cal L}|+1} \sum_{(b',c')\in {\cal L}\cup \{(b,c)\}} \nabla \log P(Y_{b'c'}|\vec{x}_{b'},\vec{w})\\
    &= \frac{1}{|{\cal L}|+1} \sum_{(b',c')\in {\cal L}} \nabla \log P(Y_{b'c'}|\vec{x}_{b'},\vec{w}) + \frac{1}{|{\cal L}|+1} \nabla \log P(Y_{bc}|\vec{X}_b,\vec{w})\\
    &= \frac{|{\cal L}|}{|{\cal L}|+1} G_{{\cal L}} + \frac{1}{|{\cal L}|+1} \nabla \log P(Y_{bc}|\vec{X}_b,\vec{w}).
    \end{aligned}
\end{equation}
We proceed with the assumption that the minimization of the marginal log-likelihood was successfully accomplished prior to the current query such that $G_{{\cal L}}|_{\vec{w}} = 0$ for the resulting $\vec{w}$. Consequently, we can simplify the expression for the gradient of marginal log-likelihood after adding the pair $(b,c)$ as:
\begin{equation}
    \begin{aligned}
    G_{{\cal L}\cup\{(b,c)\}} = \frac{1}{|{\cal L}|+1} \| \nabla \log P(Y_{bc}|\vec{X}_b,\vec{w}) \|.
    \end{aligned}
\end{equation}
Because $Y_{bc} \in \{0,1\}$, then the EGL of the marginal log-likelihood after adding pair $(b,c)$ is computed as:
\begin{equation}
        \begin{aligned}
        EGL_{bc} &= \frac{1}{|{\cal L}|+1} (P(Y_{bc}=1|\vec{X}_b,\vec{w})
        \| \nabla \log P(Y_{bc}=1|\vec{X}_b,\vec{w}) \| \\
        &+ P(Y_{bc}=0|\vec{X}_b,\vec{w})\| \nabla \log P(Y_{bc}=0|\vec{X}_b,\vec{w}) \|),
        \end{aligned}
    \end{equation}
where 
\begin{equation}
\begin{aligned}
   &\log P(Y_{bc}=Y|\vec{X}_b,\vec{w}) \\
   &=  I(Y=0) \sum_{i=1}^{n_b} \log P(y_{bi} \neq c) +I(Y=1) \log  (1 - e^{\sum_{i=1}^{n_b} \log P(y_{bi} \neq c)}) 
\end{aligned}
\end{equation}
as in (\ref{eq:mml_final_function}) and 
\begin{equation}
\begin{aligned}
    &\nabla_{\vec{w}_t} \log P(Y_{bc}=Y|\vec{X}_b,\vec{w}) \\
    &=\sum_{i=1}^{n_b}  \biggl(I(Y=1) P(Y_{bc}=0)\frac{P(y_{bi}=c)}{P(Y_{bc}=1)}-I(Y=0)P(y_{bi}=c)  \biggr)  \biggl(I(t=c) - I(t\neq c)\frac{P(y_{bi} = t)}{1-P(y_{bi} = c)}\biggr) \vec{x}_{bi}
    \end{aligned}
\end{equation}
as in (\ref{eq:grad_MML}).\\
The beg-class pair that satisfies the following condition is selected for querying:
           \begin{equation}
             (b^*,c^*) =\arg \max_{b,c} EGL_{bc}.
         \end{equation}   
Our selection based on EGL approach requires no use of the labeled data and for each bag-class pair only the features of bag $b$, $\vec{X}_b$,  and the model parameter vector $\vec{w}$ are needed. The computational complexity of the selection phase is $\textstyle{O(\sum_{b \in {\cal U}} n_b C d)}$, where as before $n_b$ is the number of instances in bag $b$, $C$ is the number of classes, and $d$ is the dimension of the feature vector $\vec{x}_{bi}$.       

\subsubsection{Uncertainty sampling for MIML} 
Uncertainty sampling is a common approach to active learning in the standard supervised setting. For probabilistic classifiers, this involves applying the classifier to each unlabeled
instance and querying those with most uncertainty about the class label. Specifically, the learner should query instance $\vec{x}$  about which it is least certain how to label. Let $P(y|\vec{x},\vec{w})$ be the probability that instance $\vec{x}$ belongs to class $y$. A more general uncertainty sampling variant might query the instance whose prediction is the least confident:
\begin{equation}
    \vec{x}^* = \arg\max_{\vec{x}\in {\cal U}} (1 - P(\hat{y}|\vec{x},\vec{w})),
\end{equation}
where $\hat{y} = \arg\max_{y\in \{1, \dots,C\}} P(y|\vec{x},\vec{w})$. This uncertainty sampling approach is applied in single-instance single-label (SISL) or multi instance learning (MIL). To apply this uncertainty sampling principle directly to multi-label learning, we need to determine $\textstyle{\hat{Y} = \arg\max_{Y\in \{0,1\}^C} P(Y|\vec{x},\vec{w})}$. The computational complexity of this step is $O(2^C)$, which leads  to the same issue of computational complexity with the aforementioned general EGL when $C$ is large. We restrict our attention to selecting the bag-class pair which (i) avoids the high computation cost associated with the calculation of the uncertainty sampling criterion for an entire bag; (ii) avoids labeling potentially-redundant classes by querying the most informative class only.
To do so, we adopt the idea of uncertainty sampling for binary classification. In this setting, uncertainty sampling queries the instance, which is nearest to the boundary. For example, in \cite{settles2008multiple}, instance is selected to query its label if this instance satisfies:
\begin{equation}
    \vec{x}^* = \arg\max_{\vec{x} \in {\cal U}} ~2 P(y=1|\vec{x},\vec{w})(1-P(y=1|\vec{x},\vec{w})),
\end{equation}
Specifically, instead of considering all classes of a bag,  we propose an approach to select a bag-class pair from unlabeled data to query based on uncertainty sampling. Given the model parameters and the current training data, for each bag-class pair in the unlabeled data, we compute its score based on the probability of the class being present in the bag. The bag-class pair that its class probability is close to the boundary is considered as the most informative pair and will be selected to add to training data for model update. \\ 
For each $(b,c) \in U$, we compute the score of each unlabeled pair as:
        \begin{equation}
            S_{bc} = 2P(Y_{bc} =1|\vec{X}_b,\vec{w}))(1-P(Y_{bc}=1|\vec{X}_b,\vec{w})),
        \end{equation}
where $P(Y_{bc} =1|\vec{X}_b,\vec{w}) = 1- \prod_{i=1}^{n_b} p(y_{bi} \neq c)$ and $P(Y_{bc} =0) = 1 - P(Y_{bc} =1) $.
Note that we can also compute the score using the following formula $S_{bc} = |P(Y_{bc} =1 |\vec{X}_b,\vec{w})-0.5\|P(Y_{bc}=0|\vec{X}_b,\vec{w})-0.5|$. The pair that satisfies the following condition is selected for querying:
           \begin{equation}
             (b^*,c^*) =\arg \max_{b,c} S_{bc}.
         \end{equation} 
 As with EGL, our selection based on uncertainty sampling approach requires  only the features of bag $b$, $\vec{X}_b$,  and the model parameter vector $\vec{w}$. The computational complexity of the selection phase is $\textstyle{O(\textstyle{\sum_{b \in {\cal U}} n_b C d)}}$, where $n_b$ is the number of instances in bag $b$, $C$ is the number of classes, and $d$ is the dimension of the feature vector $\vec{x}_{bi}$.
\subsection{Model Update} \label{sec:model_update}
 In this section, we present the model update process after every query. We will begin with the update of training data after one query is performed and proceed with the update of the model parameters.
\subsubsection{Training data update}
After querying the bag-class pair from the unlabeled pool, we update this pair into the training data. Specifically, let $b^{*k},c^{*k}$ is the selected pair at the  $k$th query, $t \in \{0, 1\}$ is the value of $Y_{b^{*k}c^{*k}}$, and $S_1^k, \dots,S_B^k $ are the available label set of $B$ bags at the $k$th query, we have:
\begin{equation}\label{eq:updatequery}
		\begin{aligned}
		\nonumber  
         Y_{bc}^{k}=
        \begin{cases}
        Y_{bc}^{k-1}, ~~ \text{ if }, (b,c)\neq (b^{*k},c^{*k}) \\
        t, ~~ \text{ if}, (b,c) = (b^{*k},c^{*k}) .
        \end{cases}
		\end{aligned}
		\end{equation}
and
\begin{equation}\label{eq:updatequery_sb}
		\begin{aligned}
		\nonumber  
         S_{b}^{k}=
        \begin{cases}
        S_b^{k-1}, ~~ \text{ if }, b\neq b^{*k} \\
        S_b^{k-1} \cup \{c^{*k}\}, ~~ \text{ if}, b = b^{*k} ;
        \end{cases}
		\end{aligned}
		\end{equation}
\begin{equation}
    {\cal U} \leftarrow {\cal U} \backslash \{(b^{*k},c^{*k})\}
\end{equation}
\subsubsection{Model  parameter update}\label{subsec:MMLupdate}
We demonstrate how our model is updated after a bag-class pair is queried. Specifically, After each query, our goal is to update the model parameters by minimizing the MML objective that takes into account all the available labels:
\begin{equation}
\begin{aligned}
	\bar{F}^k(\vec{w}) =  \frac{1}{\sum_b^B|S_b^k|}\sum_{i=1}^k f_{b^{*i}c^{*i}}(\vec{w},Y_{b^*c^*}^i),
\end{aligned}
\end{equation}
where 
\begin{equation}\label{eq:f_bc}
    f_{bc}(\vec{w},l) =  -[I(l=0) \sum_{i=1}^{n_b} \log P(y_{bi} \neq c) + I(l=1) \log  (1 - e^{\sum_{i=1}^{n_b} \log P(y_{bi} \neq c)})]
\end{equation}
is the objective associated with the label of bag $b$ and class $c$. 
To reduce the computational complexity associated with the minimization of the MML objective that includes the entire available data, we propose a stochastic gradient descent (SGD) approach that involves updating the parameter vector based on the gradient associated with the newly obtained label:
\begin{equation}\label{eq:update_w}
    \begin{aligned}
    \vec{w}^k &= {\cal P}(\vec{w}^{k-1} - \eta_k g^k(\vec{w}^{k-1})),
    \end{aligned}
\end{equation}
where $g^k = \nabla f_{b^{*k}c^{*k}}(\vec{w},Y_{b^{*k}c^{*k}})$ is the gradient of $f_{b^{*k}c^{*k}}(\vec{w},Y_{b^*c^*}^k)$, $\eta_k$ is the step-size, and ${\cal P}$ is used to denote a projection onto the feasible solution set. For the gradient, 
 $g^k$ is computed by
\begin{equation}\label{eq:pair}
\begin{aligned}
g^k=-\sum_{i=1}^{n_{b_k}} & \biggl(
I(Y_{b_kc_k}=1) P(Y_{b_kc_k}=0)\frac{P(y_{b_ki}=c_k)}{P(Y_{b_kc_k}=1)}- I(Y_{b_kc_k}=0)P(y_{b_ki}=c_k)  \biggr) \\
&  \biggl(I(c=c_k) - I(c\neq c_k)\frac{P(y_{b_ki} = c)}{1-P(y_{b_ki} = c_k)}\biggr) \vec{x}_{b_ki}.
\end{aligned}
\end{equation}
We consider a monotonically decreasing step size that follows this form: $\textstyle{\eta_k =\frac{c'}{\lambda k + c''}}$, where $c'$ and $c''$ are constants, $\lambda$ is the regularization term. 
Similar to algorithms such as PEGASOS \cite{shalev2011pegasos}, we can show that the optimal solution is guaranteed to be contained in a sphere of a given radius. For our problem, we can show that the radius is $\textstyle{\tau = \sqrt{\frac{2}{\lambda}\max(\log(C),\max_b (n_b) \frac{1}{C-1})}}$ and the solution must reside in the sphere ${\cal S}_{\tau}=\{\vec{w}~|~\|\vec{w}\| \le \tau \}$. The projection operator onto the sphere ${\cal S}_{\tau}$ is denoted by ${\cal P}$ and is give by
\[
{\cal P}(\vec{w})= \left\{ \begin{array}{c c c} 
\vec{w}, & \quad & \|\vec{w}\| \le \tau,\\
\tau \frac{\vec{w}}{\| \vec{w} \|}, & \quad & \|\vec{w}\| > \tau. 
\end{array} \right.
\]
The detailed derivation of $\tau$ is provided in Appendix~\ref{appendix:bound_norm}.\\
Consider the MML as the average of bag-class pairs to solve the minimization problem using SGD leads to more randomness and increases the variant of the gradient. Therefore, we introduce another method for update the model parameter after a bag-label pair comes. Instead of update only the queried pair, we update the full bag that contain the bag-label queried pair. Specifically, after each query, we update the model parameters by minimizing the following MML: 
\begin{equation}
\begin{aligned}
	\bar{F}^k(\vec{w}) =  \frac{1}{\sum_b^B|S_b^k|}\sum_{b=1}^B \sum_{c\in S_b^k}f_{bc}(\vec{w},Y_{bc}^k),
\end{aligned}
\end{equation}
where $f_{bc}(\vec{w},Y_{bc}^k)$ is computed as in (\ref{eq:f_bc}). 
The update rule is the same with the bag-class update (\ref{eq:update_w}):
\begin{equation}\label{eq:update_w1}
    \begin{aligned}
    \vec{w}^k &= {\cal P}(\vec{w}^{k-1} - \eta_k g^k(\vec{w}^{k-1})),
    \end{aligned}
\end{equation}
where $\eta_k$ and ${\cal P}(\cdot)$ are described right after (\ref{eq:update_w}). The difference here is that $g^k$ is the gradient of the full bag
 $\textstyle{g^k = \nabla \sum_{c_k \in S_{b_k}}f_{b_kc_k}(\vec{w},Y_{b_kc_k})}$ and $g^k$ w.r.t $\vec{w}_c$. This bag contains the queried bag-class pair. The gradient $g^k$ is computed as:
\begin{equation}\label{eq:bag}
\begin{aligned}
g^k=-\sum_{c_k \in S_{b_k}}\sum_{i=1}^{n_{b_k}} & \biggl(
I(Y_{b_kc_k}=1) P(Y_{b_kc_k}=0)\frac{P(y_{b_ki}=c_k)}{P(Y_{b_kc_k}=1)}- I(Y_{b_kc_k}=0)P(y_{b_ki}=c_k)  \biggr) \\
&  \biggl(I(c=c_k) - I(c\neq c_k)\frac{P(y_{b_ki} = c)}{1-P(y_{b_ki} = c_k)}\biggr) \vec{x}_{b_ki}.
\end{aligned}
\end{equation}
From (\ref{eq:pair}) and (\ref{eq:bag}), the only difference between the bag-class pair update and the full bag update is the gradient of the marginal log-likelihood $g^k$ used in updating the model parameters. In the bag-class pair update, $g^k$ is the gradient of the bag-class pair queried. Other available class labels of this bag are not taken into account to update the model. However, in the full bag update, all other available class labels beside the queried bag-class pair are used in updating the model.
\section{Experiments}
In this section, we evaluate our proposed approach for learning an instance/bag-level classifier under the MIML active learning with missing labels by deploying three different experimental settings: (i) model update inference, we compare our online (SGD) approach with offline GD approach in \cite{nguyen2020incomplete} to verify the correctness of our approach; (ii) instance selection criterion and type comparison: we run our model with four different selection criteria; specifically, we compare our proposed bag-class pair selection based on EGL and uncertainty sampling with other two selection criteria: bag selection and bag-then-label selection from \cite{retz2016active} and \cite{huang2017multi}, respectively, to verify the effectiveness of our selection criterion and type; (iii) finally, we compare our framework with two other state-of-the-art approaches on the MIML setting with active learning \cite{retz2016active,huang2017multi}. To perform the comparison, with each compared method, we run our model with the instance selection strategy used by the compared method. In our comparison, we evaluate bag-level  metrics as a function of the epochs and the number of bag-class pair queries. \\
\textbf{Dataset:} We perform a comparison on three benchmark datasets, including: HJA - a bird song audio recordings dataset \cite{briggs2012acoustic}, 
and two letter datasets
\cite{briggs2012rank} (i.e., Letter Carol and Letter Frost). HJA is bird song audio recordings dataset. Each $10$-second audio recording is converted into a bag consisting of audio segments obtained via time-frequency domain segmentation and featurized as in \cite{briggs2012acoustic}. The bags are manually labeled to indicate the presence or absence of bird species. HJA dataset includes $645$ bags with a total of $13$ bird species. Letter Carol and Letter Frost are also MIML datasets, each is taken from a poem \cite{briggs2012rank}. Each word (bag) contains multiple letters. Each letter is described by a $16$-dimensional feature vector and is annotated by one of $26$ letter labels\footnote{Specifically at this point, $24$ letters are present in poem} from `a' to `z'. The labels for each word are the union of its letter labels. On each dataset, we generate $10$ cross validation sets and the results are reported based on the average of these $10$ sets.
\subsection{Model update inference methods - comparison}
In this section, we run our online SGD and offline GD approach in \cite{nguyen2020incomplete} on three aforementioned datasets. In our online SGD, we run experiments when model is updated with new coming bag, called bag-SGD and with new coming bag-class pair, called pair-SGD. For each dataset and on each cross validation set, we run all training data to learn the model parameter. We run three algorithms until convergence to show that the performance of our online SGD approach is comparable to GD approach. \\
\textbf{Evaluation metrics:} We report the results based on bag accuracy used for MIML learning evaluation in \cite{tsoumakas2009mining} as the function of the number of epochs. Besides, we present the log-likelihood function as the function of number of epochs.\\
\begin{figure*}[htb]
\begin{subfigure}{0.33\linewidth}
    \resizebox{\linewidth}{!}{\includegraphics[]{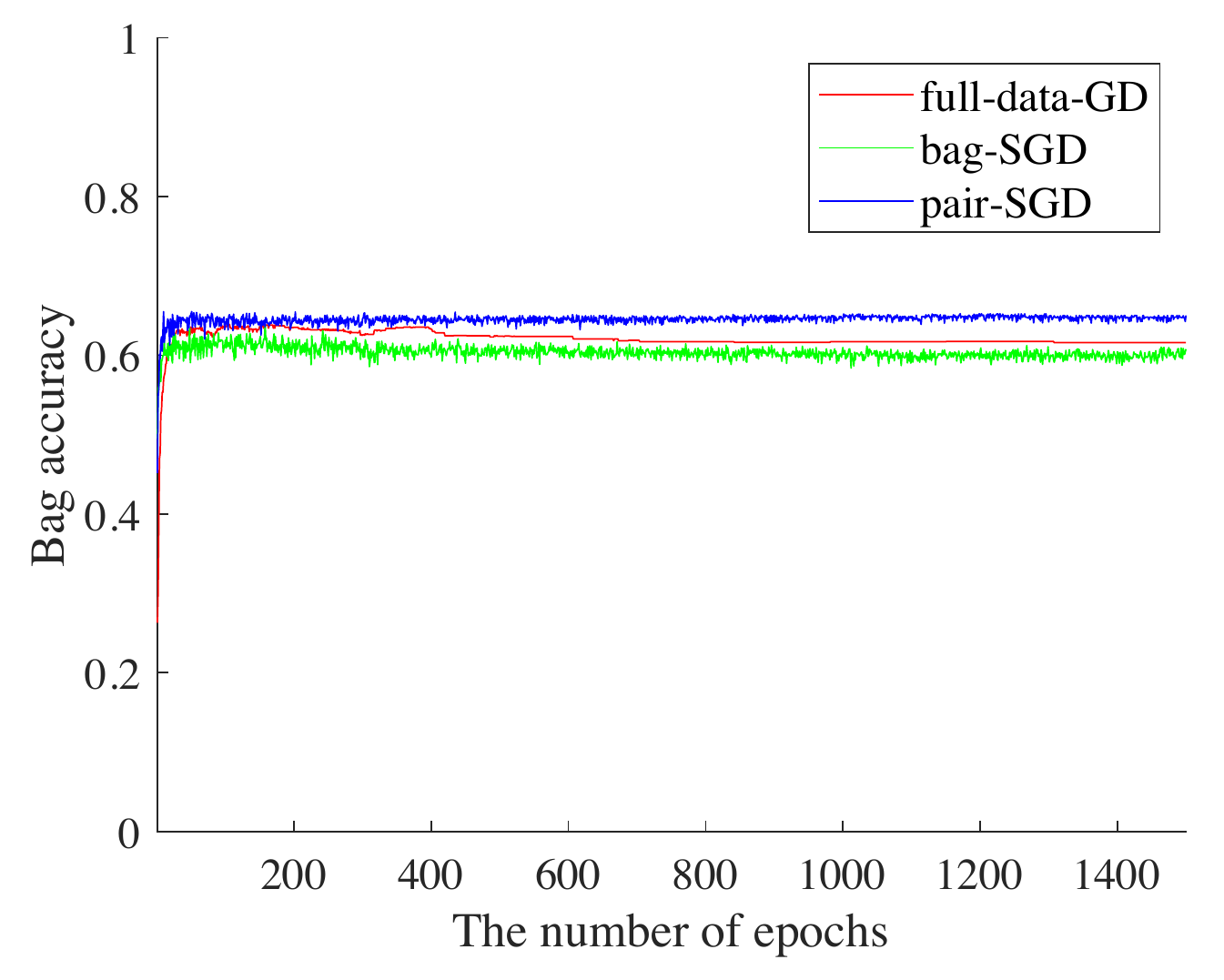}}
    \caption{Carroll}
\end{subfigure}
\begin{subfigure}{0.33\linewidth}
   \resizebox{\linewidth}{!}{\includegraphics[]{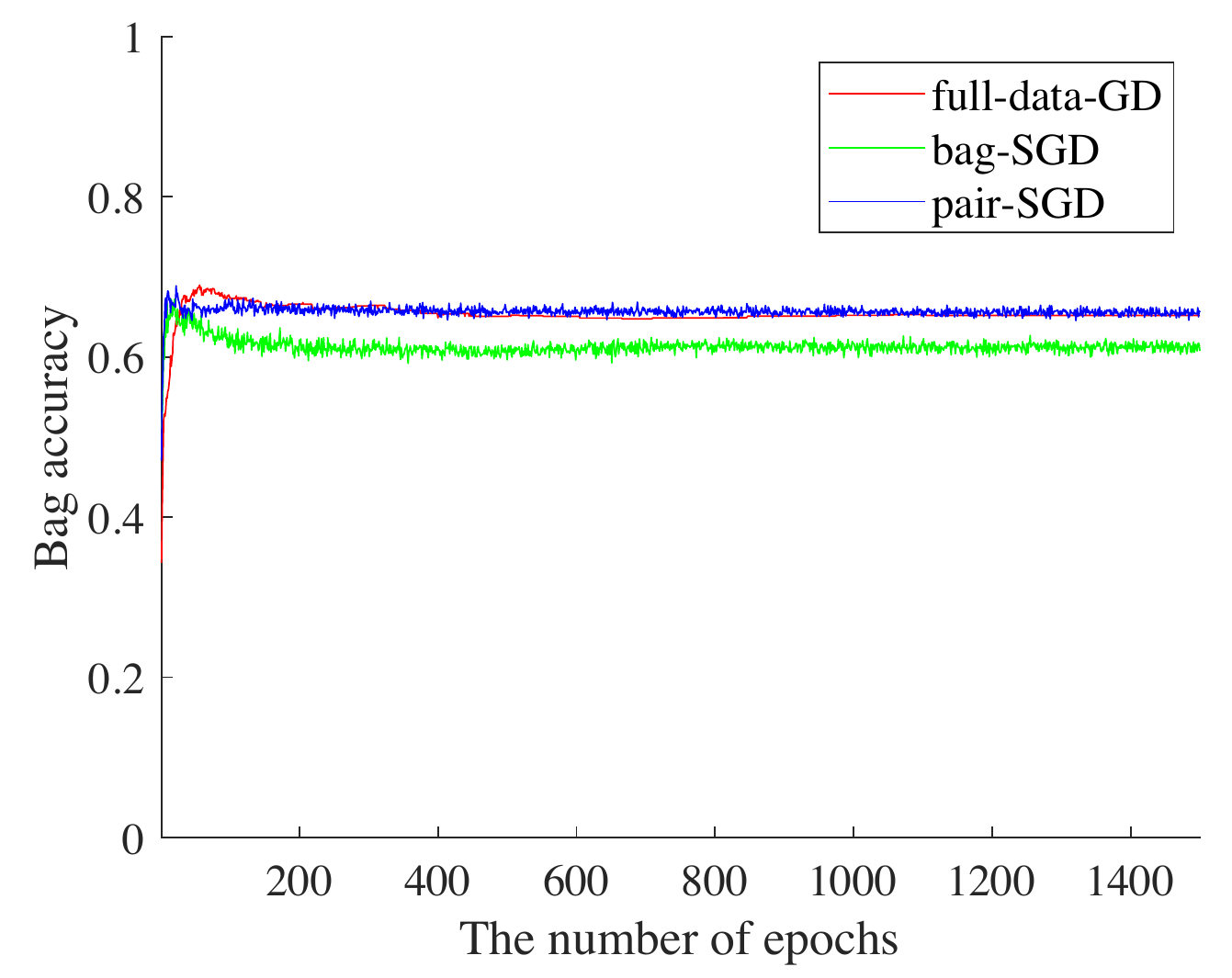}}
    \caption{Frost}
\end{subfigure}
\begin{subfigure}{0.33\linewidth}
   \resizebox{\linewidth}{!}{\includegraphics[]{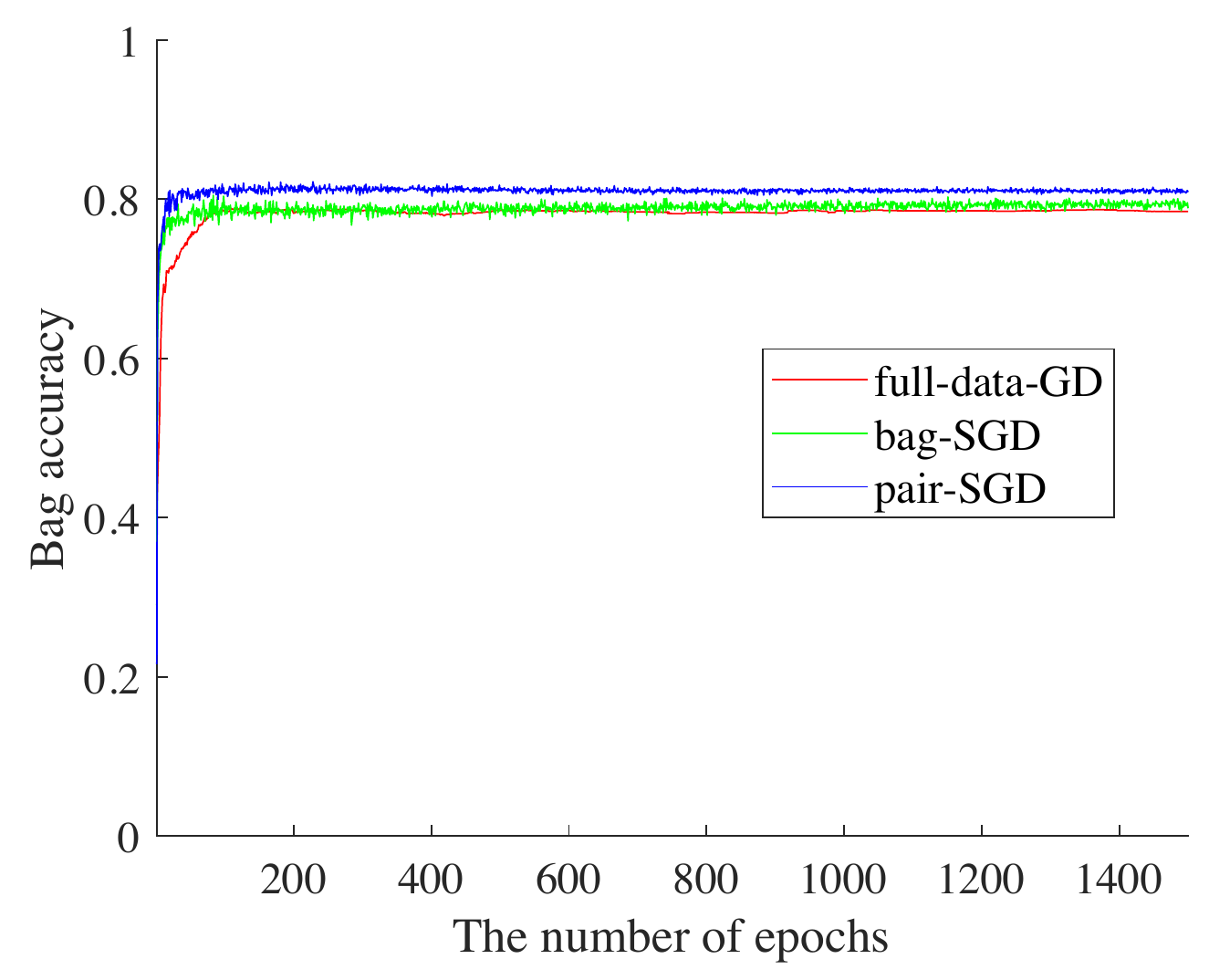}}
    \caption{HJA}
\end{subfigure}
\caption{Bag accuracy as the function of the number of epochs of three model update inference methods: full-data GD, bag-SGD and pair-SGD on three datasets: Carroll, Frost, and HJA.} \label{fig:bagaccuracy_online_offline}\vspace{-.5em}
\end{figure*} 
\begin{figure*}[htb]
\begin{subfigure}{0.33\linewidth}
   \resizebox{\linewidth}{!}{\includegraphics[]{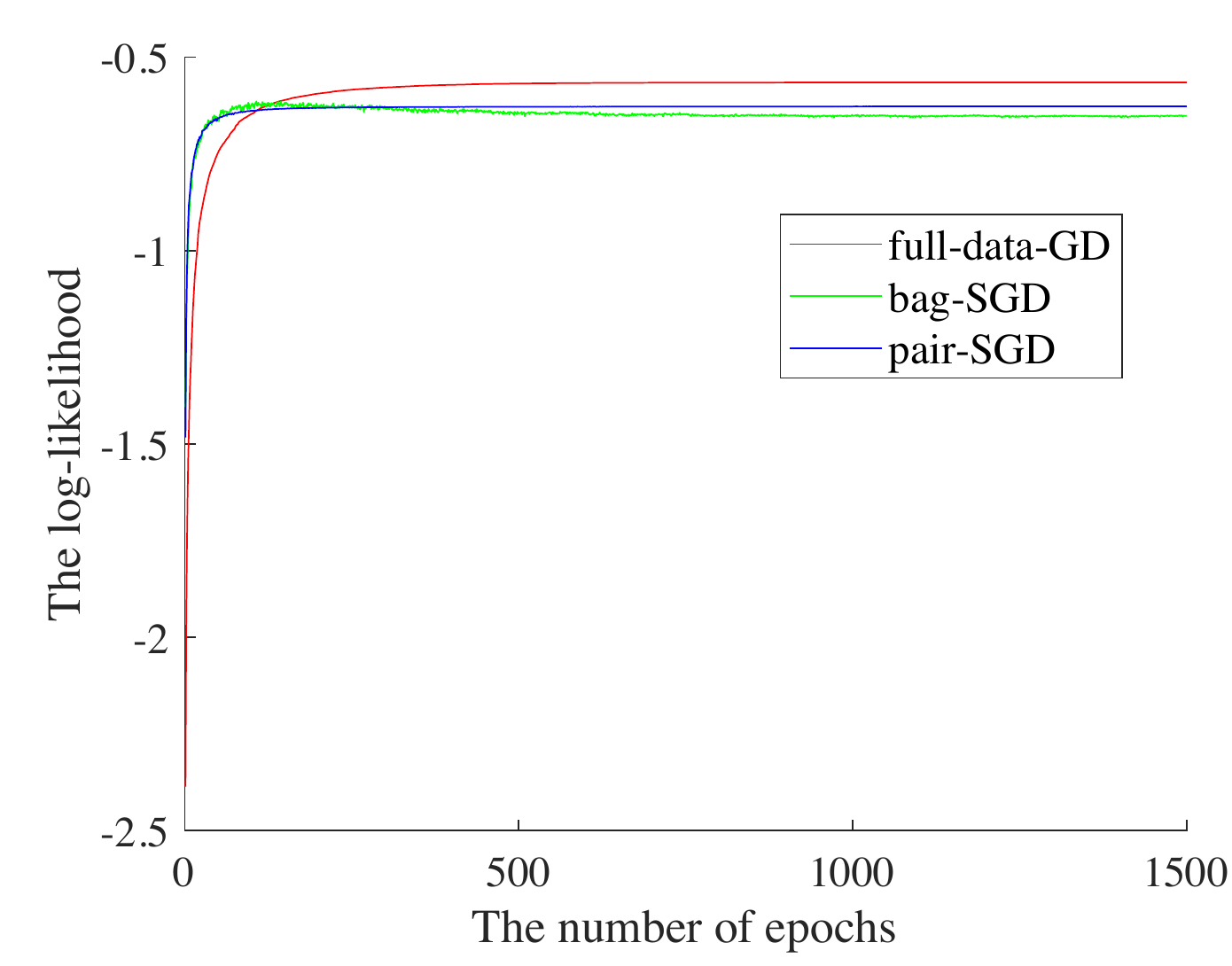}}
    \caption{Carroll}
\end{subfigure}
\begin{subfigure}{0.33\linewidth}
    \resizebox{\linewidth}{!}{\includegraphics[]{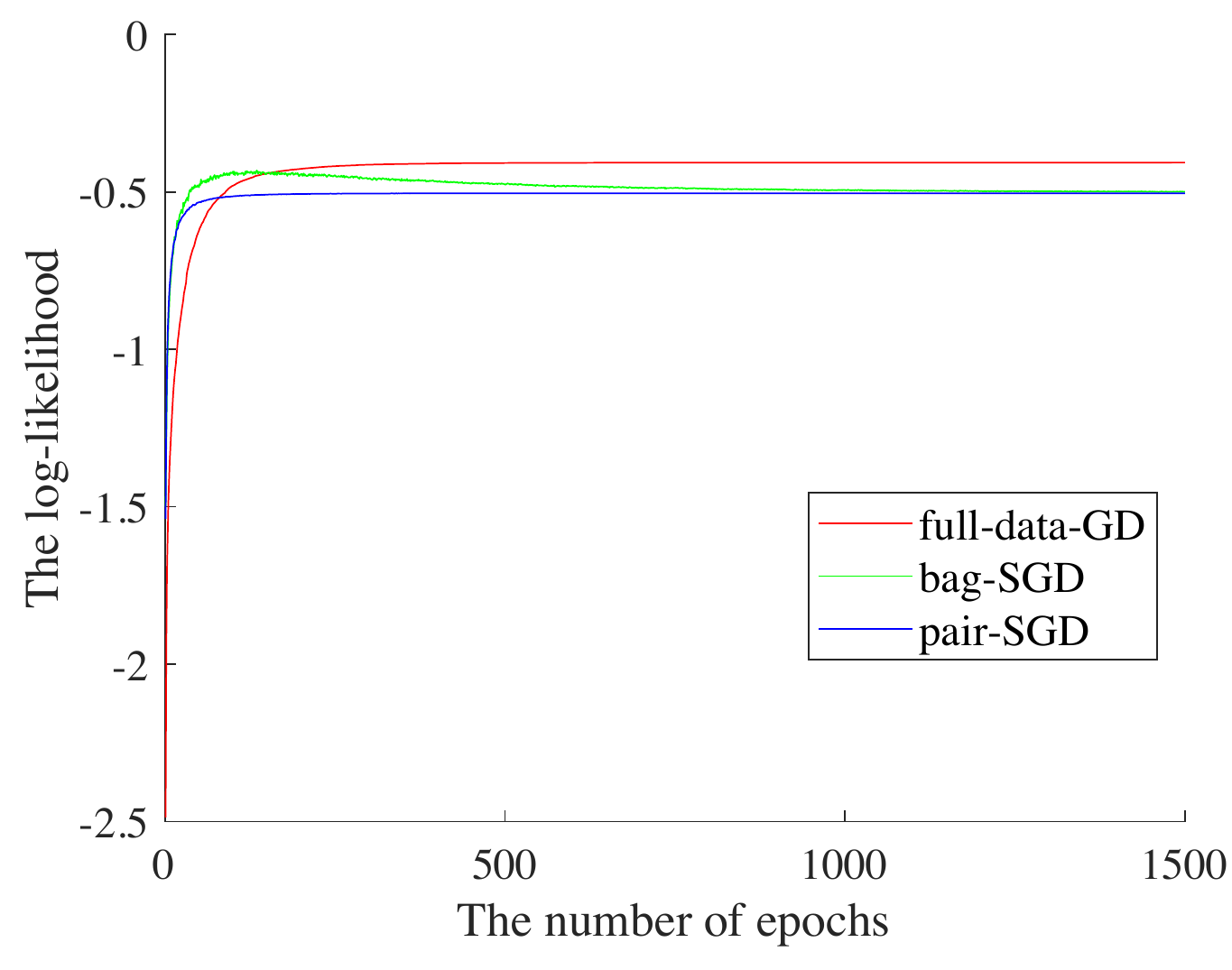}}
    \caption{Frost}
\end{subfigure}
\begin{subfigure}{0.33\linewidth}
    \resizebox{\linewidth}{!}{\includegraphics[]{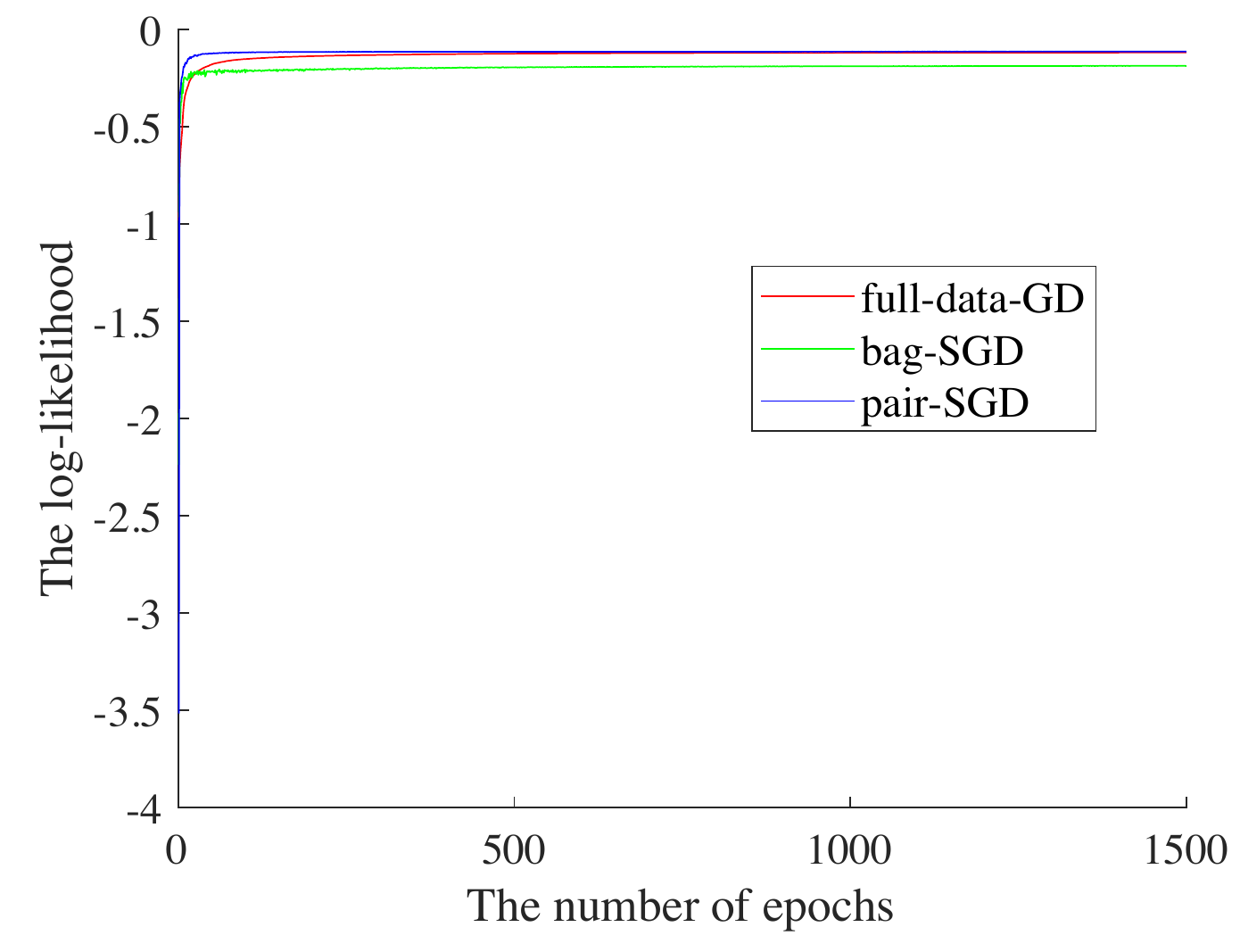}}
    \caption{HJA}
\end{subfigure}
\caption{ Log-likelihood as the function of the number of epochs of three model update inference methods: full-data GD, bag-SGD and pair-SGD on three datasets: Carroll, Frost, and HJA. 
} \label{fig:llh_online_offline}
\end{figure*}
\textbf{Result analysis:} The performance of our two online SGD update method (the bag-class pair update and the full bag update) vs. the full data GD update method in terms of bag accuracy and the log-likelihood is shown in Fig.~\ref{fig:bagaccuracy_online_offline} and Fig.~\ref{fig:llh_online_offline}. In these two figures, the x-axis is the number of epochs - each epoch is the number of iteration in which all training data is learned to update the model parameters. As can be seen in Fig.~\ref{fig:llh_online_offline}, the value of the log-likelihood of three methods converges after $1500$ epochs. The full data GD update takes more time to converge in comparing to the bag-SGD update and the bag-class pair update, but it converges to a higher value of log-likelihood than our two online update methods, which is expected. 
\subsection{Instance selection criterion and type}
We run our model (MIMLILL-AL) on four selection strategies selected from a combinations of two instance selection criteria: (i) EGL and (ii) uncertainty sampling and three types of query level: (1) bag only, (2) bag-class pair (our approach), and (3) bag-then-label. For (1) bag only criterion,  a bag is selected from the unlabeled data based on uncertainty criteria, then all labels of the selected bag are queried to be added into the current training data to update the model. For (2) a bag-class pair criterion (our approach), a bag-class pair is selected simultaneously. For (3) bag-then-label criterion, a bag is selected from the unlabeled data based on the uncertainty sampling and diversity, then a class of the bag(s) is queried based on the distance from the label to the thresholding dummy label. From the six combinations, we ignore two bag only selection strategies based on (i) EGL and (ii) uncertainty sampling to avoid the  computational complexity issue mentioned in Section \ref{sec:ins_selection}. To initialize all active learning methods, for each dataset and cross validation, a small number of bags is fully labeled and is used for training the initial model and the remaining bags are unlabeled. Queries are then made to label bags or bag-class pairs from the unlabeled data. Additionally, we consider random instance selection of bag-class pairs as a baseline method.\\
\textbf{Evaluation metrics:}
 We report the results based on metrics used for MIML learning evaluation in \cite{tsoumakas2009mining} including: bag accuracy, average precision, Hamming loss, and one-error as the function of the number of bag-class pair queries.\\
\begin{figure*}[htb]
\begin{subfigure}{0.33\linewidth}
    \resizebox{\linewidth}{!}{\includegraphics[]{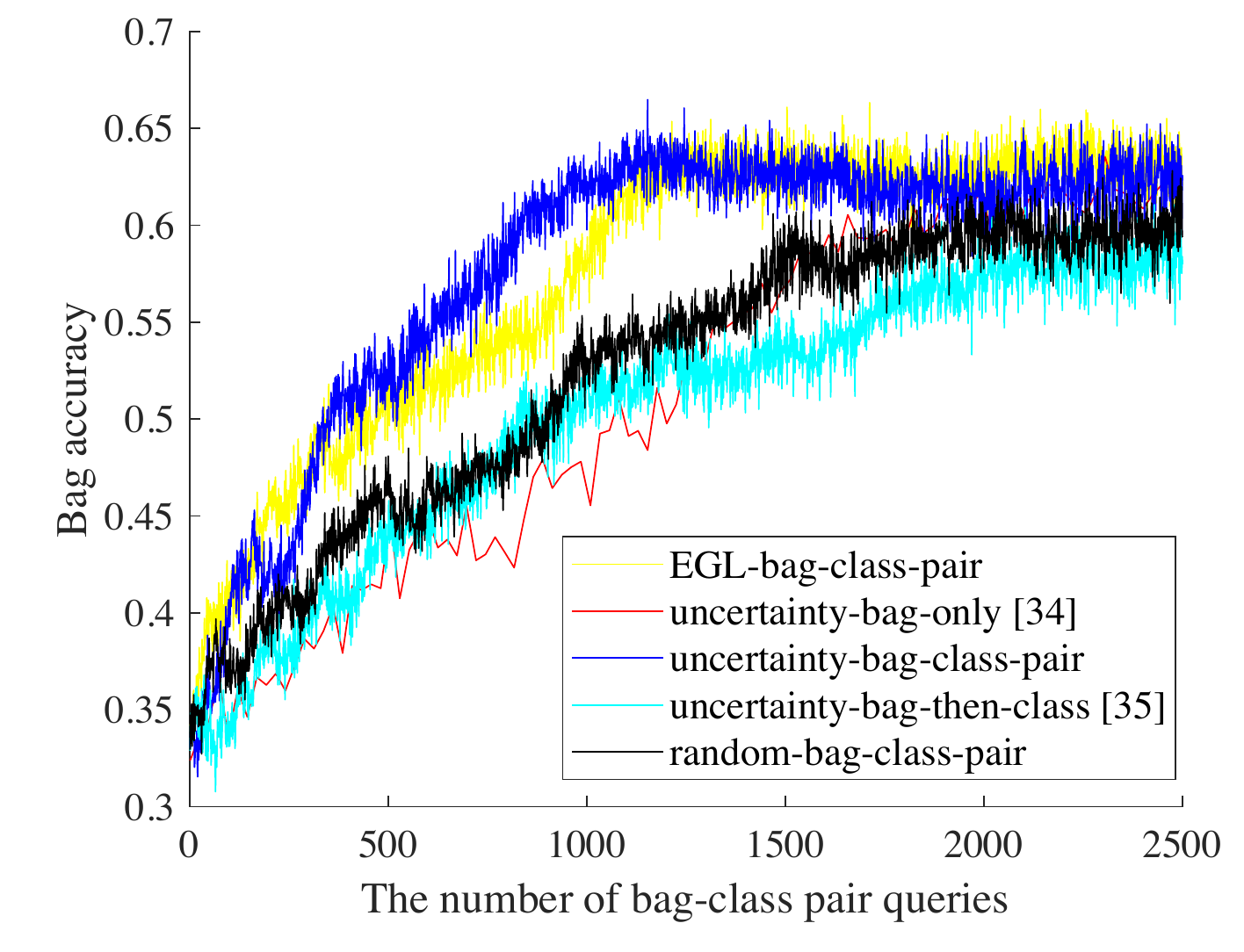}}
    \caption{Carroll}
\end{subfigure}
\begin{subfigure}{0.33\linewidth}
    \resizebox{\linewidth}{!}{\includegraphics[]{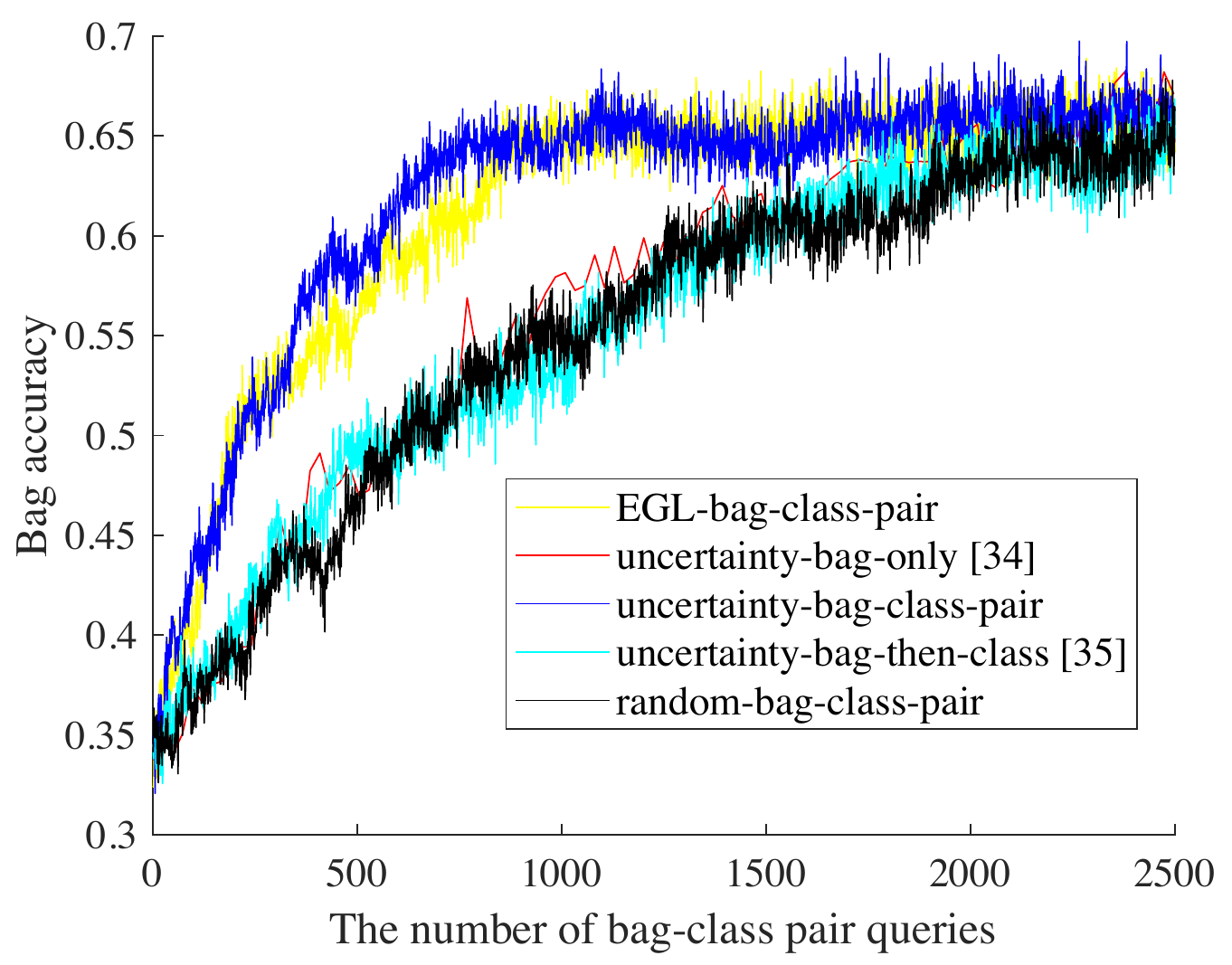}}
    \caption{Frost}
\end{subfigure}
\begin{subfigure}{0.33\linewidth}
    \resizebox{\linewidth}{!}{\includegraphics[]{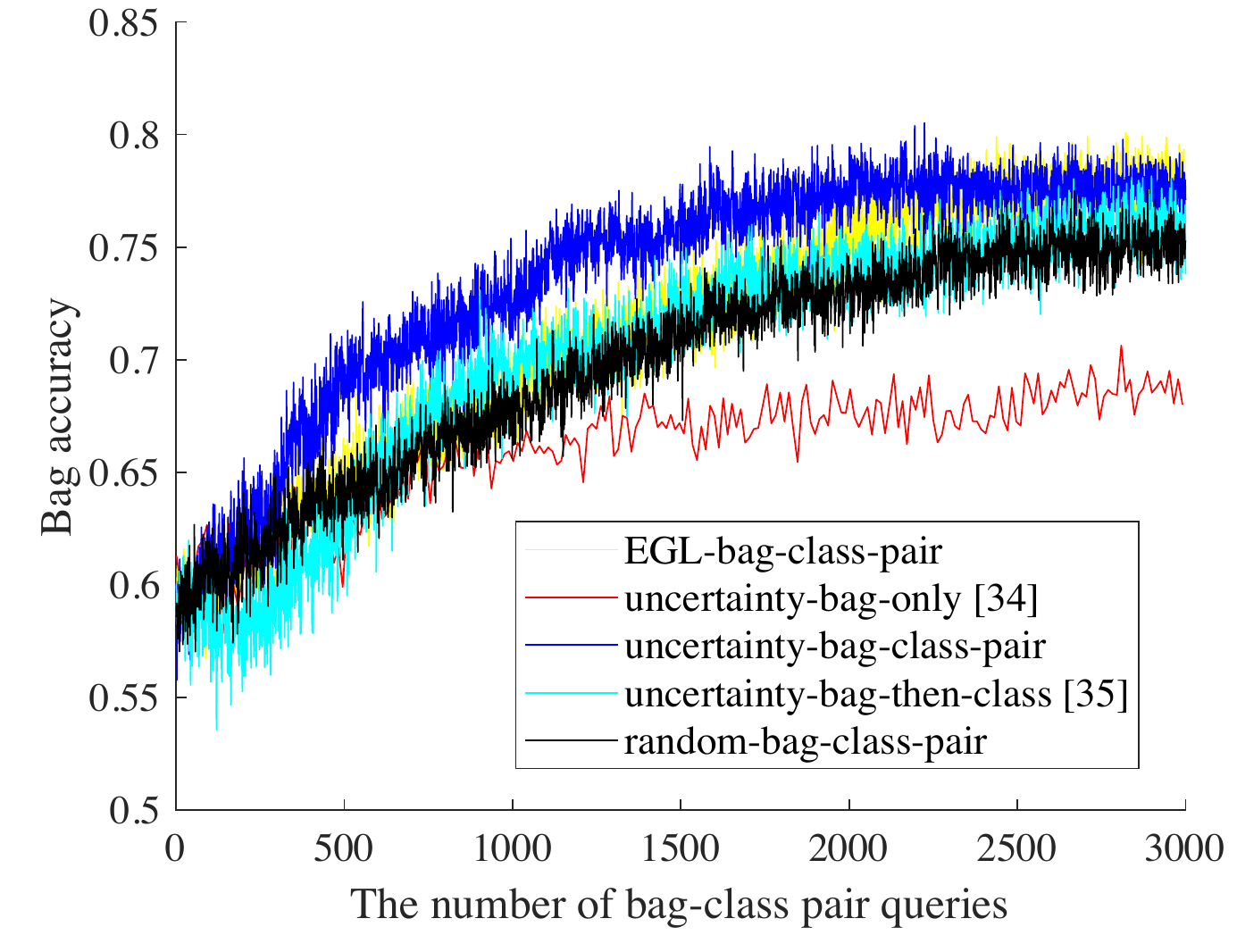}}
    \caption{HJA}
\end{subfigure}
\caption{ Bag accuracy as the function of the number of bag-class pair queries of five selection criteria on three datasets: Carroll, Frost, and HJA.} \label{fig:bag_accuracy_different_criterion}\vspace{-.5em}
\end{figure*}
\begin{figure*}[htb]
\begin{subfigure}{0.33\linewidth}
    \resizebox{\linewidth}{!}{\includegraphics[]{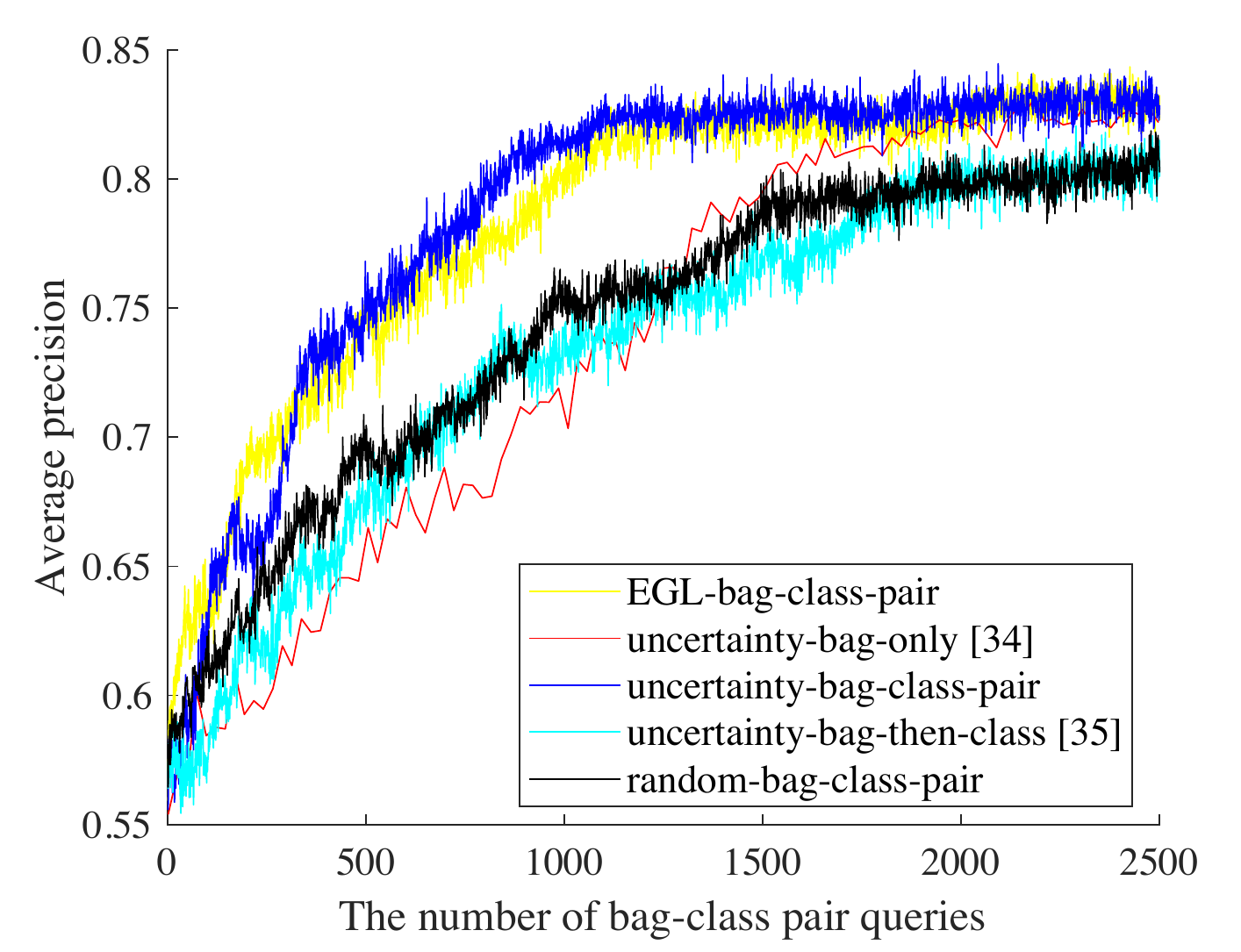}}
    \caption{Carroll}
\end{subfigure}
\begin{subfigure}{0.33\linewidth}
    \resizebox{\linewidth}{!}{\includegraphics[]{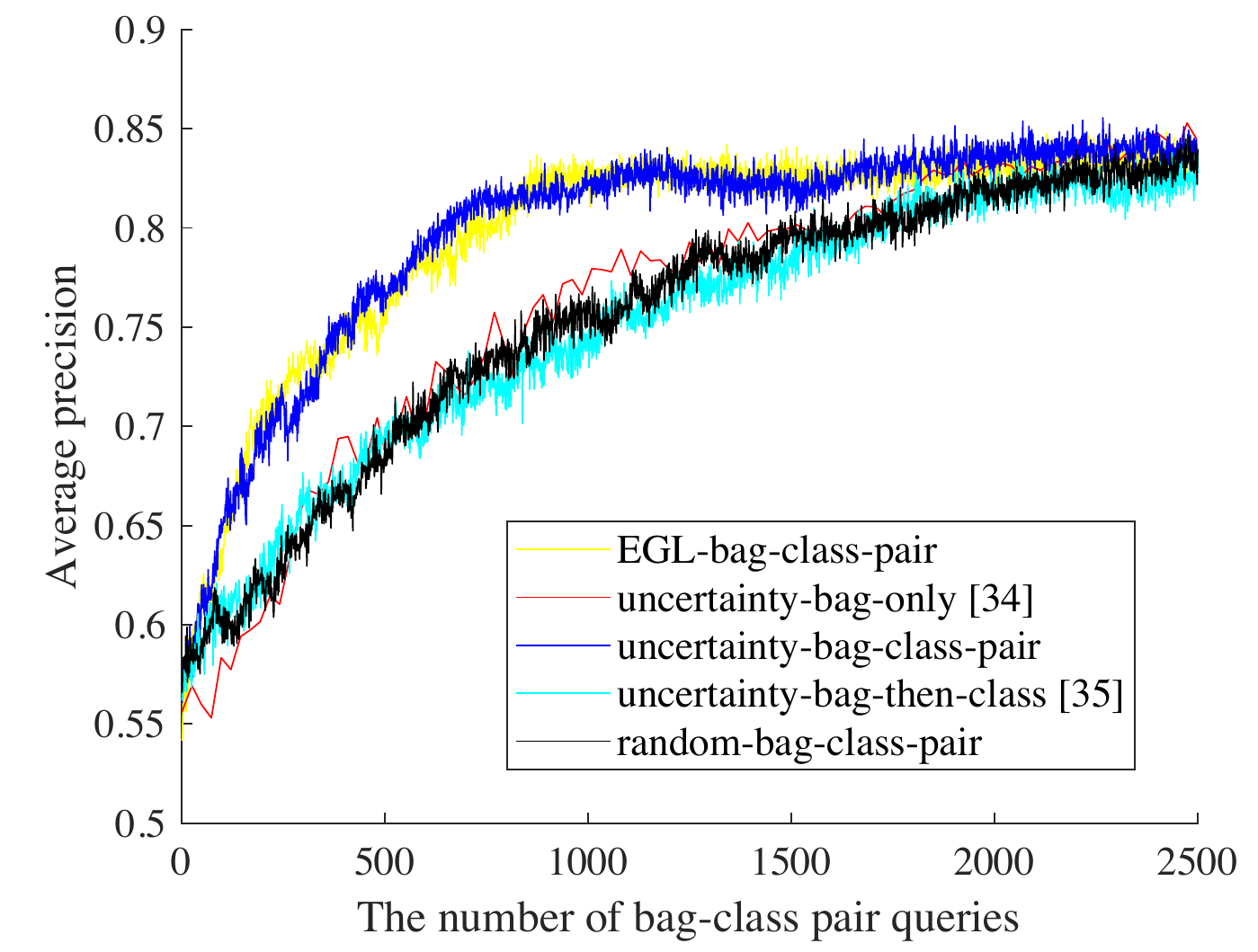}}
    \caption{Frost}
\end{subfigure}
\begin{subfigure}{0.33\linewidth}
    \resizebox{\linewidth}{!}{\includegraphics[]{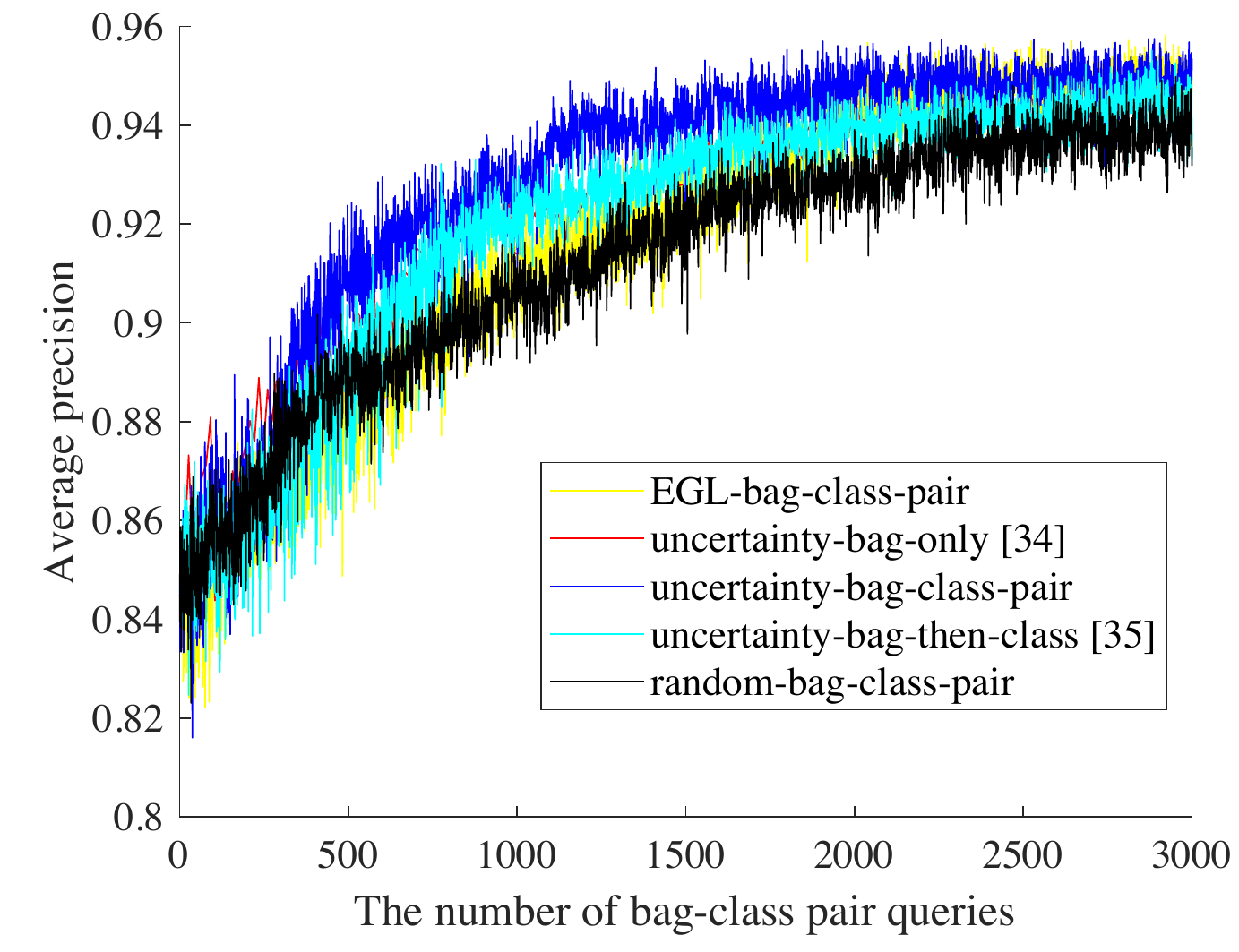}}
    \caption{HJA}
\end{subfigure}
\caption{ Average precision as the function of the number of bag-class pair queries of five selection criteria on three datasets: Carroll, Frost, and HJA.} \label{fig:avepre_different_criterion}\vspace{-.5em}
\end{figure*}
\begin{figure*}[t!]
\begin{subfigure}{0.33\linewidth}
    \resizebox{\linewidth}{!}{\includegraphics[]{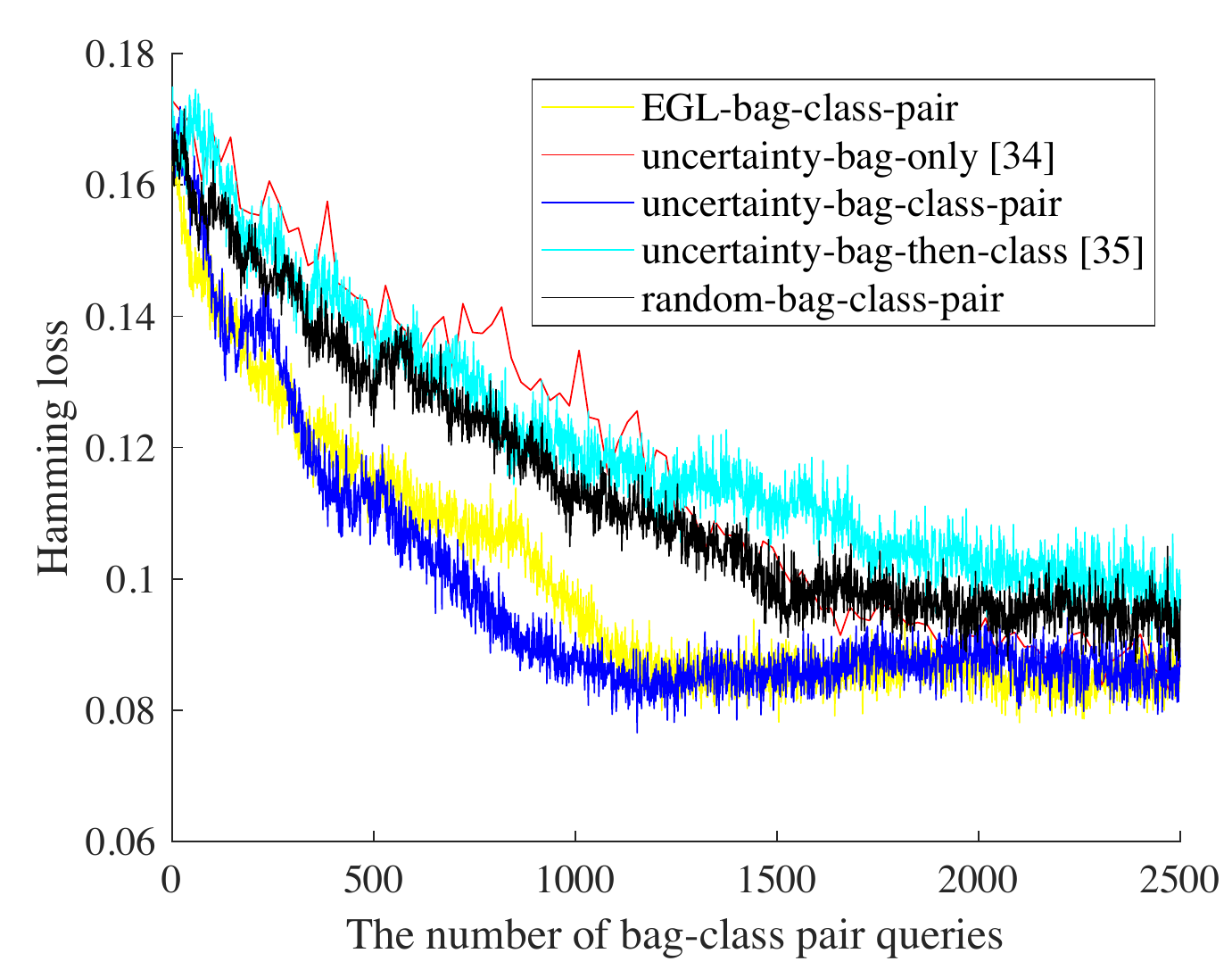}}
    \caption{Carroll}
\end{subfigure}
\begin{subfigure}{0.33\linewidth}
    \resizebox{\linewidth}{!}{\includegraphics[]{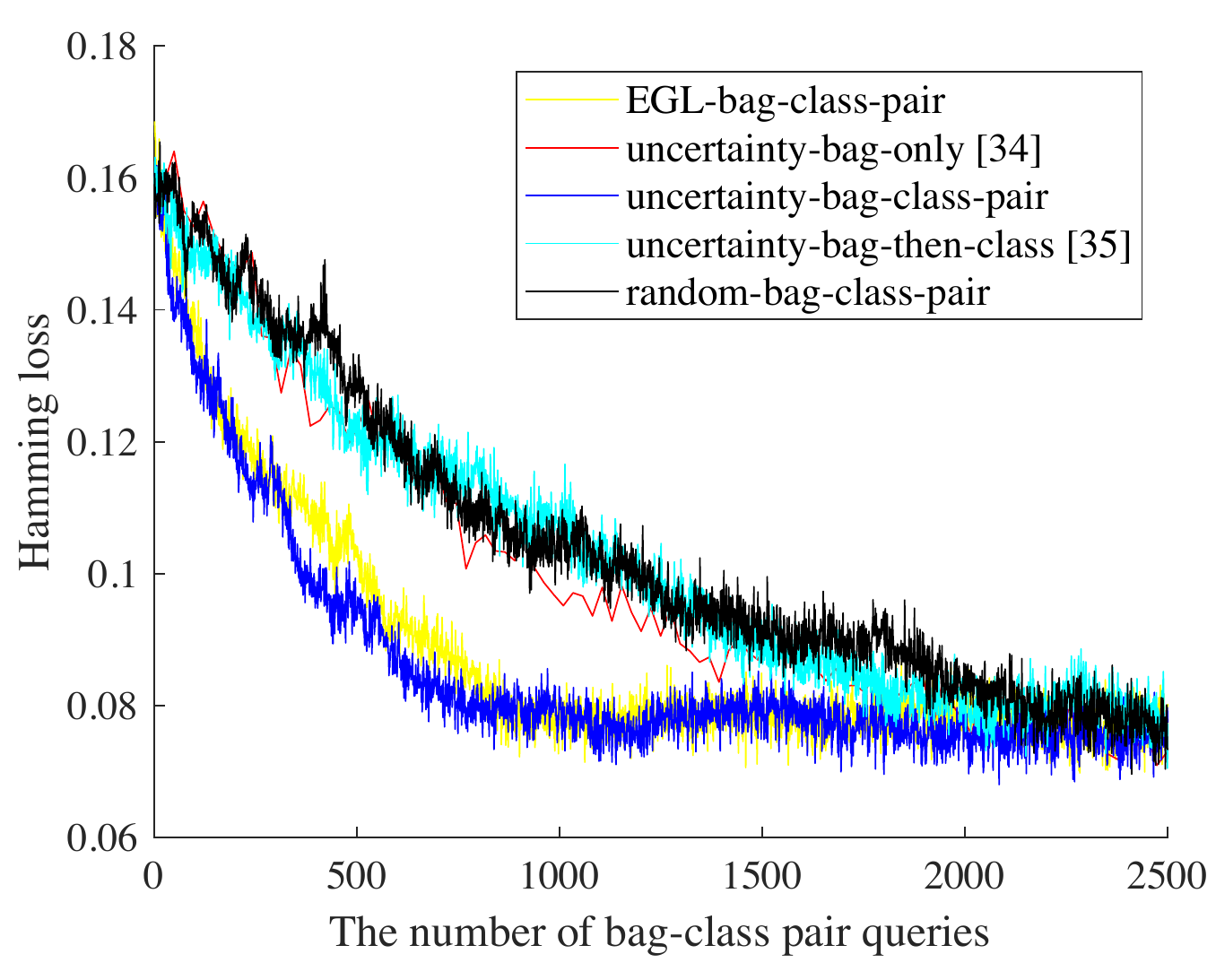}}
    \caption{Frost}
\end{subfigure}
\begin{subfigure}{0.33\linewidth}
    \resizebox{\linewidth}{!}{\includegraphics[]{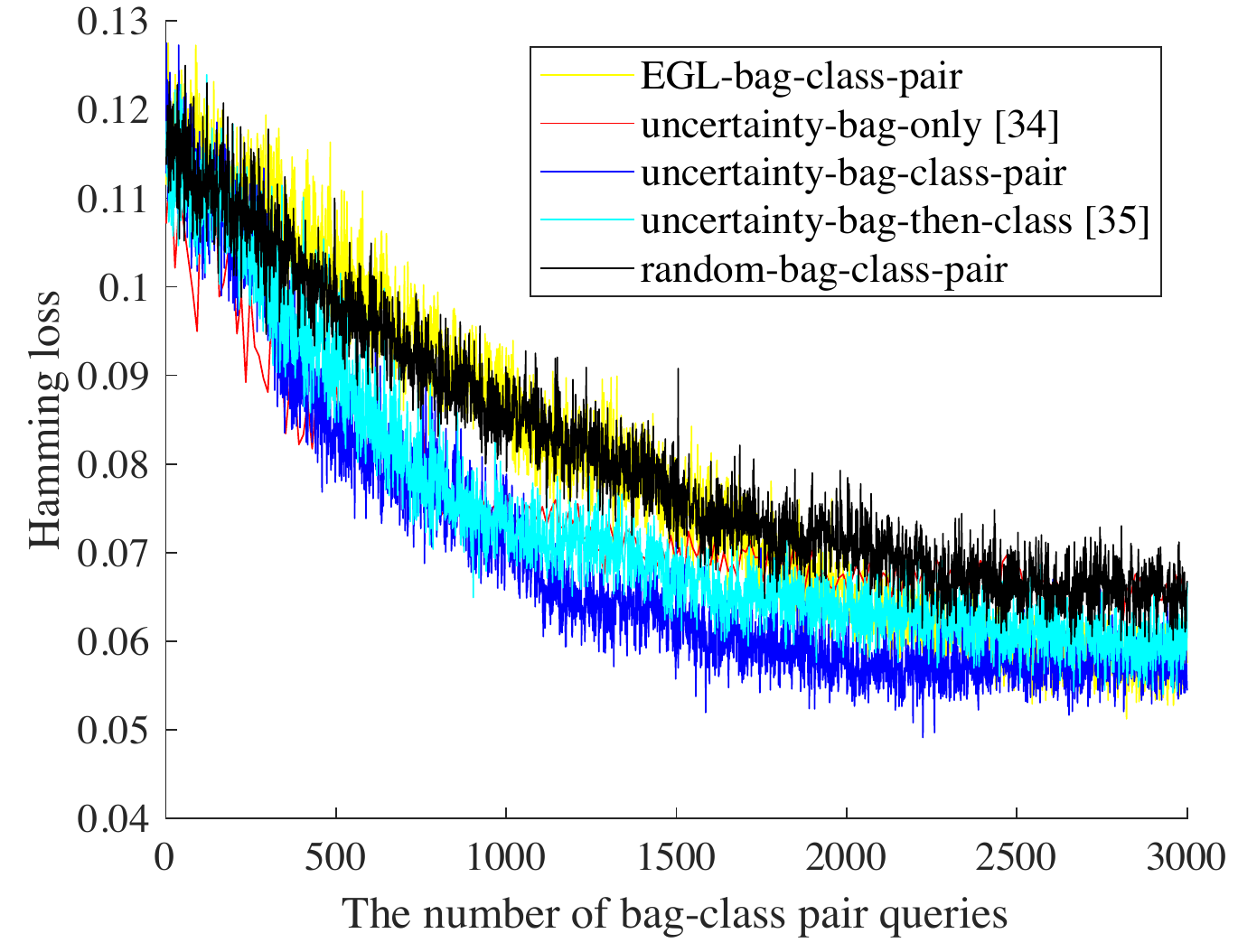}}
    \caption{HJA}
\end{subfigure}
\caption{ Hamming loss as the function of the number of bag-class pair queries of five selection criteria on three datasets: Carroll, Frost, and HJA.} \label{fig:hammingloss_different_criterion}\vspace{-.5em}
\end{figure*}
\begin{figure*}[t!]
\begin{subfigure}{0.33\linewidth}
    \resizebox{\linewidth}{!}{\includegraphics[]{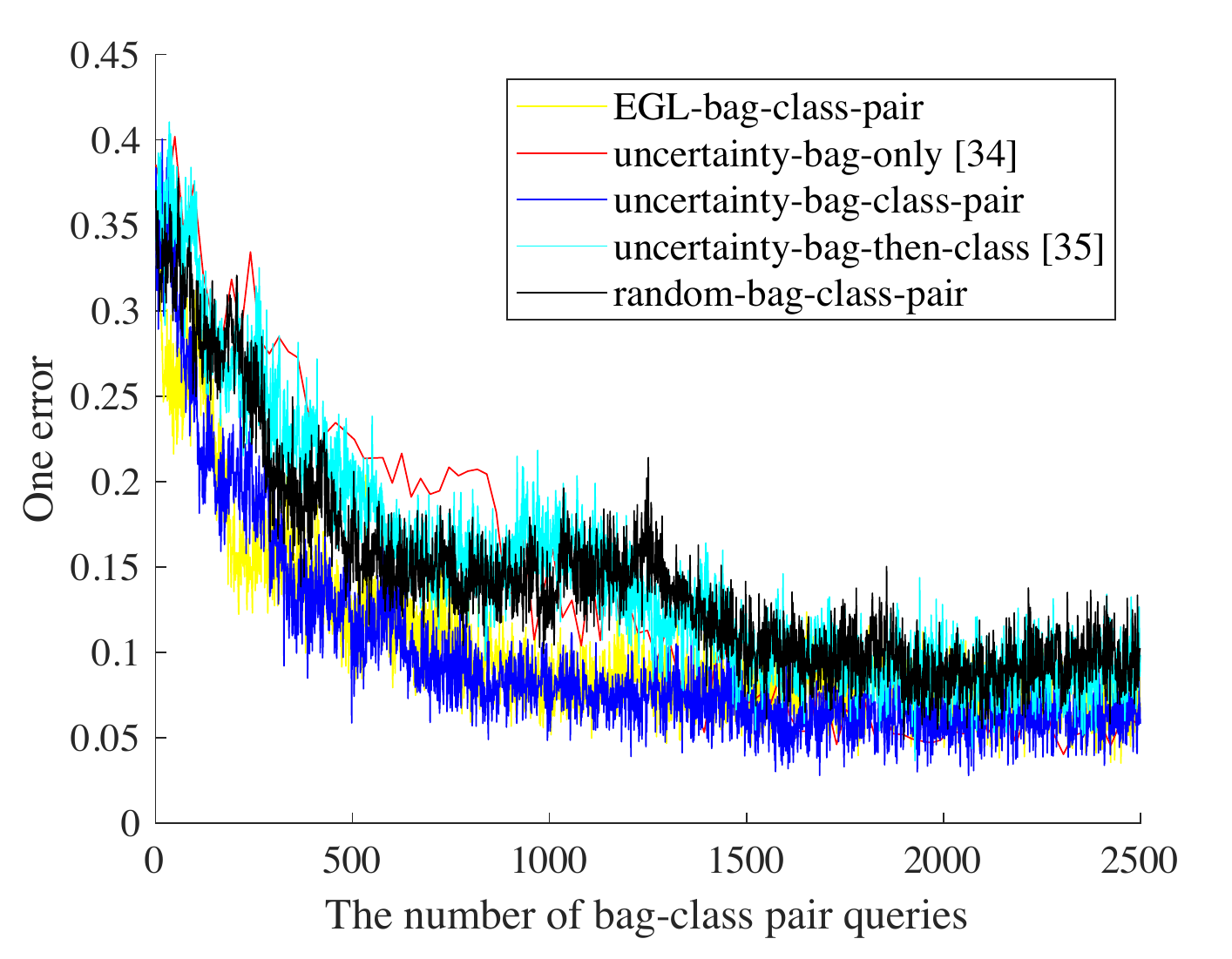}}
    \caption{Carroll}
\end{subfigure}
\begin{subfigure}{0.33\linewidth}
     \resizebox{\linewidth}{!}{\includegraphics[]{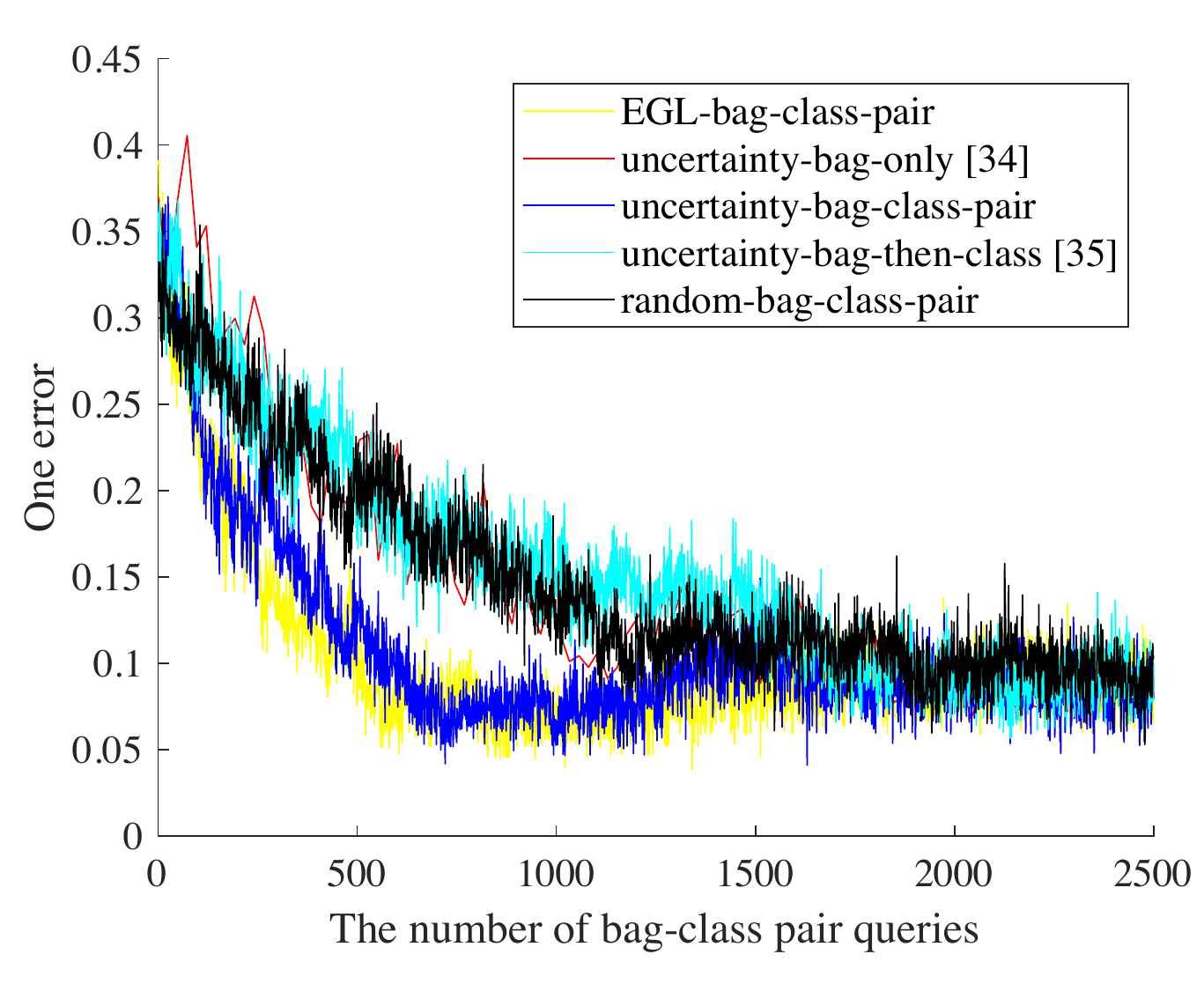}}
    \caption{Frost}
 \end{subfigure}
 \begin{subfigure}{0.33\linewidth}
     \resizebox{\linewidth}{!}{\includegraphics[]{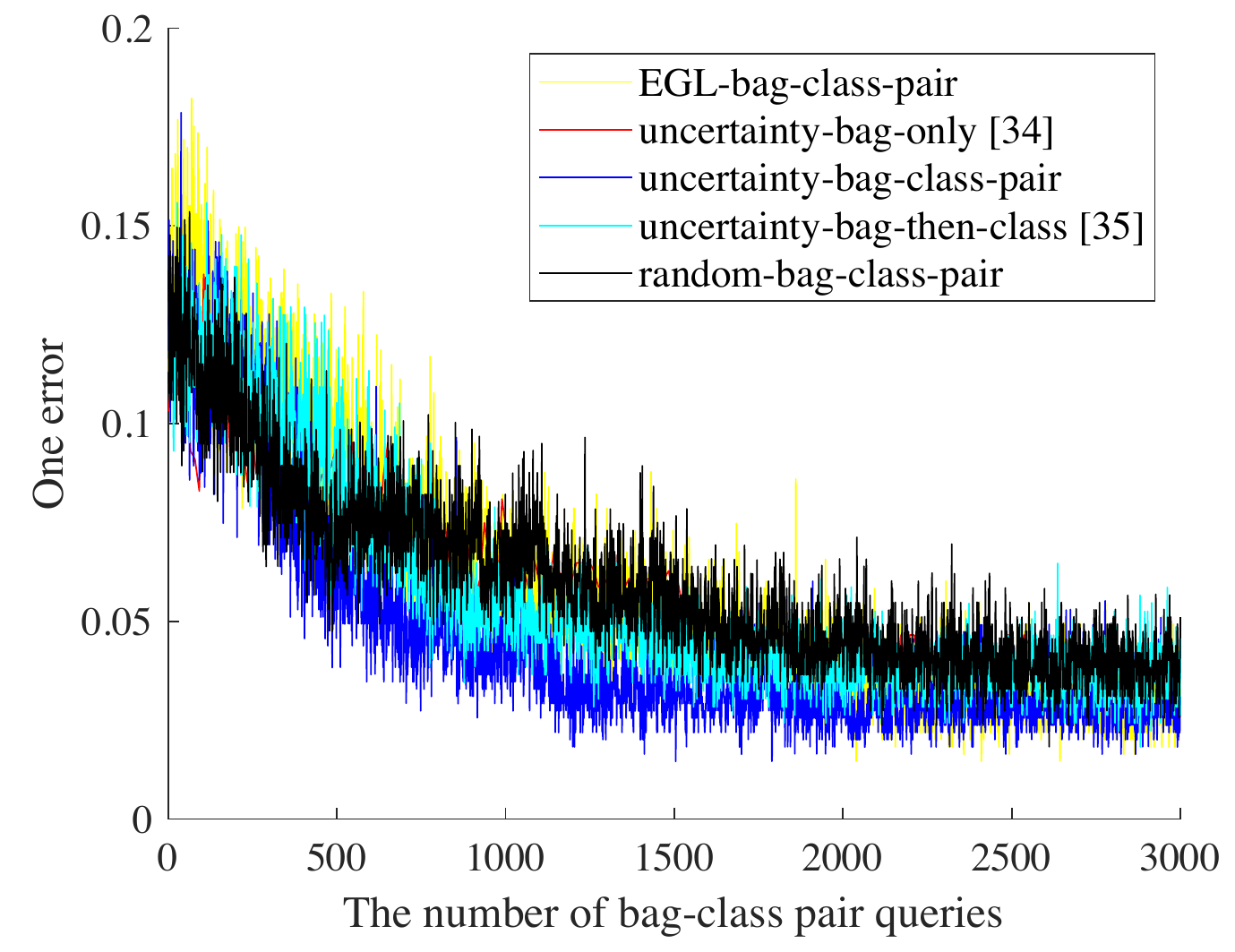}}
    \caption{HJA}
 \end{subfigure}
\caption{ One error as the function of the number of bag-class pair queries of five selection criteria on three datasets: Carroll, Frost, and HJA.} \label{fig:oe_different_criterion}\vspace{-.5em}
\end{figure*}
\begin{figure*}[t!]
\begin{subfigure}{0.33\linewidth}
    \resizebox{\linewidth}{!}{\includegraphics[]{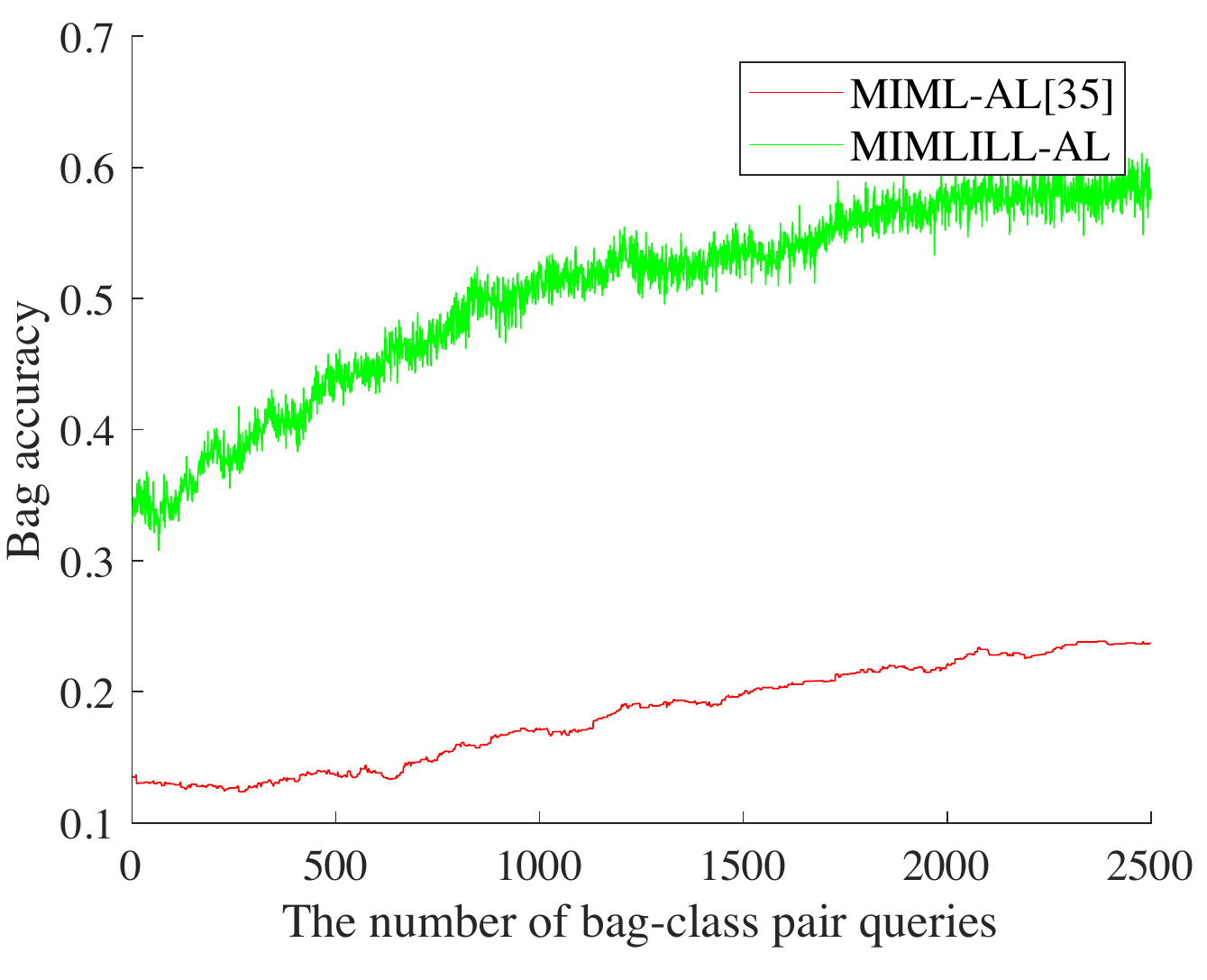}}
    \caption{Carroll}
\end{subfigure}
\begin{subfigure}{0.33\linewidth}
    \resizebox{\linewidth}{!}{\includegraphics[]{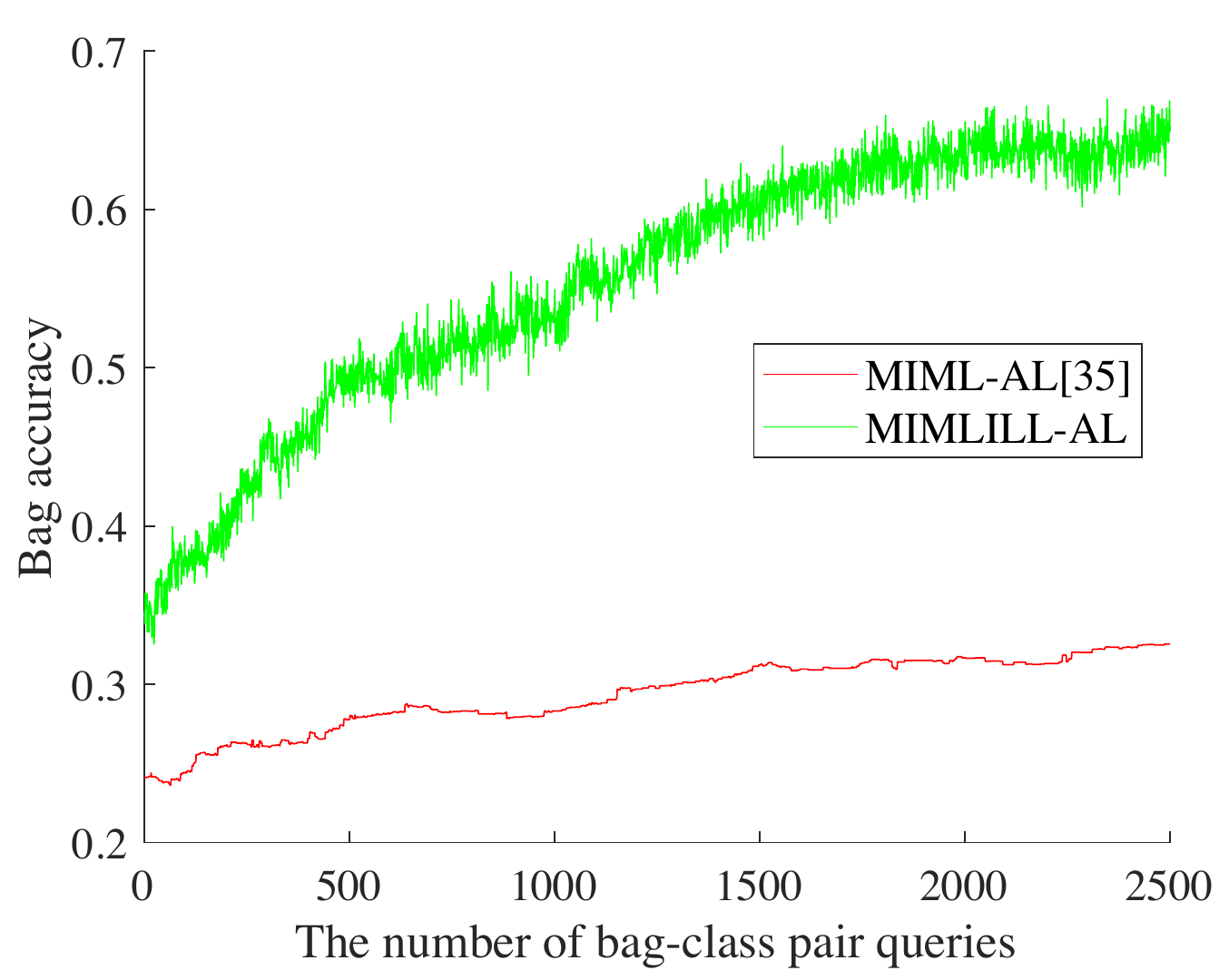}}
    \caption{Frost}
\end{subfigure}
\begin{subfigure}{0.33\linewidth}
    \resizebox{\linewidth}{!}{\includegraphics[]{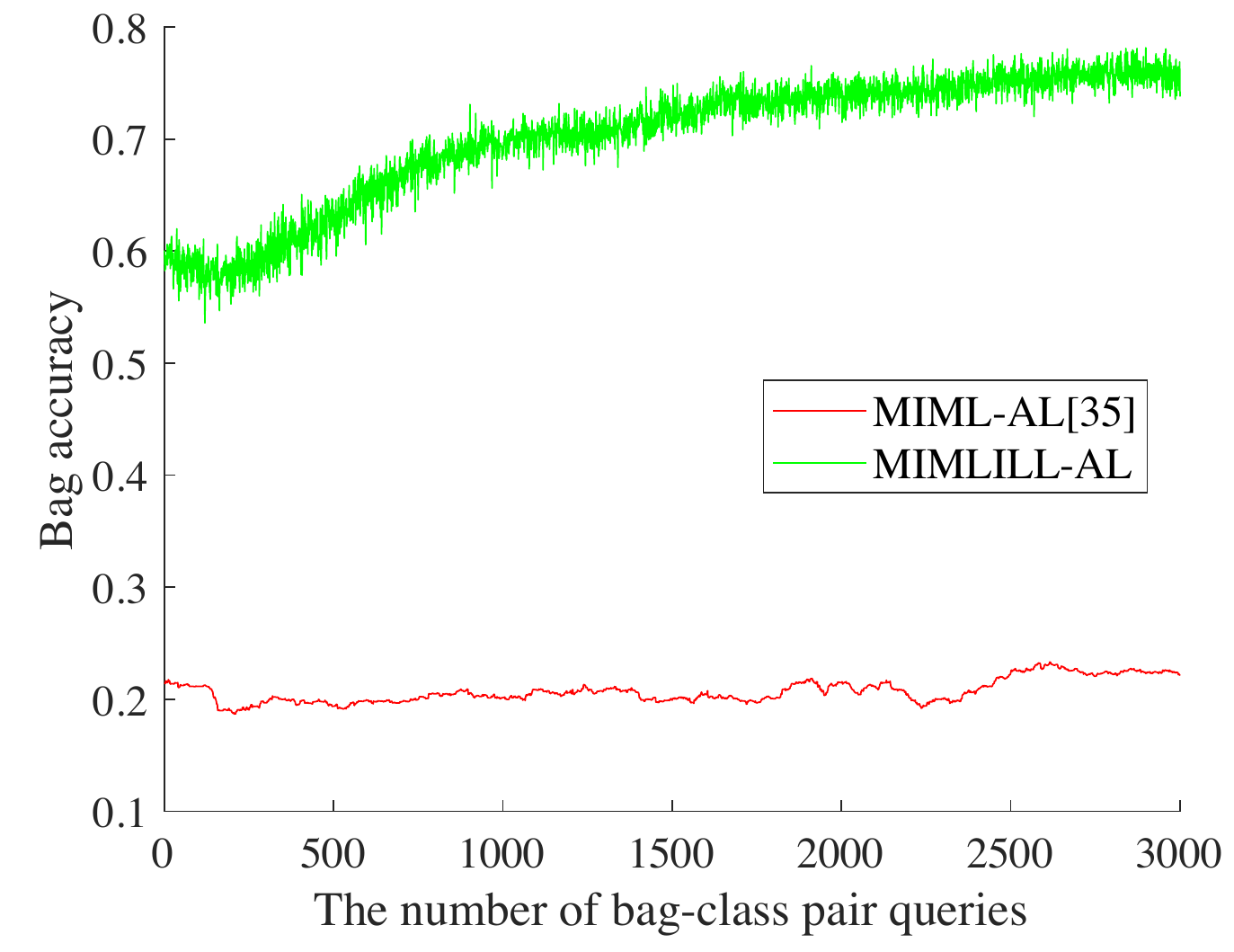}}
    \caption{HJA}
\end{subfigure}
\caption{ Bag accuracy as the function of the number of bag-class pair queries of our method and MIML-AL \cite{huang2017multi} on three datasets: Carroll, Frost, and HJA.} \label{fig:ba_vs_huang}\vspace{-.5em}
\end{figure*}
\begin{figure*}[t!]
\begin{subfigure}{0.33\linewidth}
    \resizebox{\linewidth}{!}{\includegraphics[]{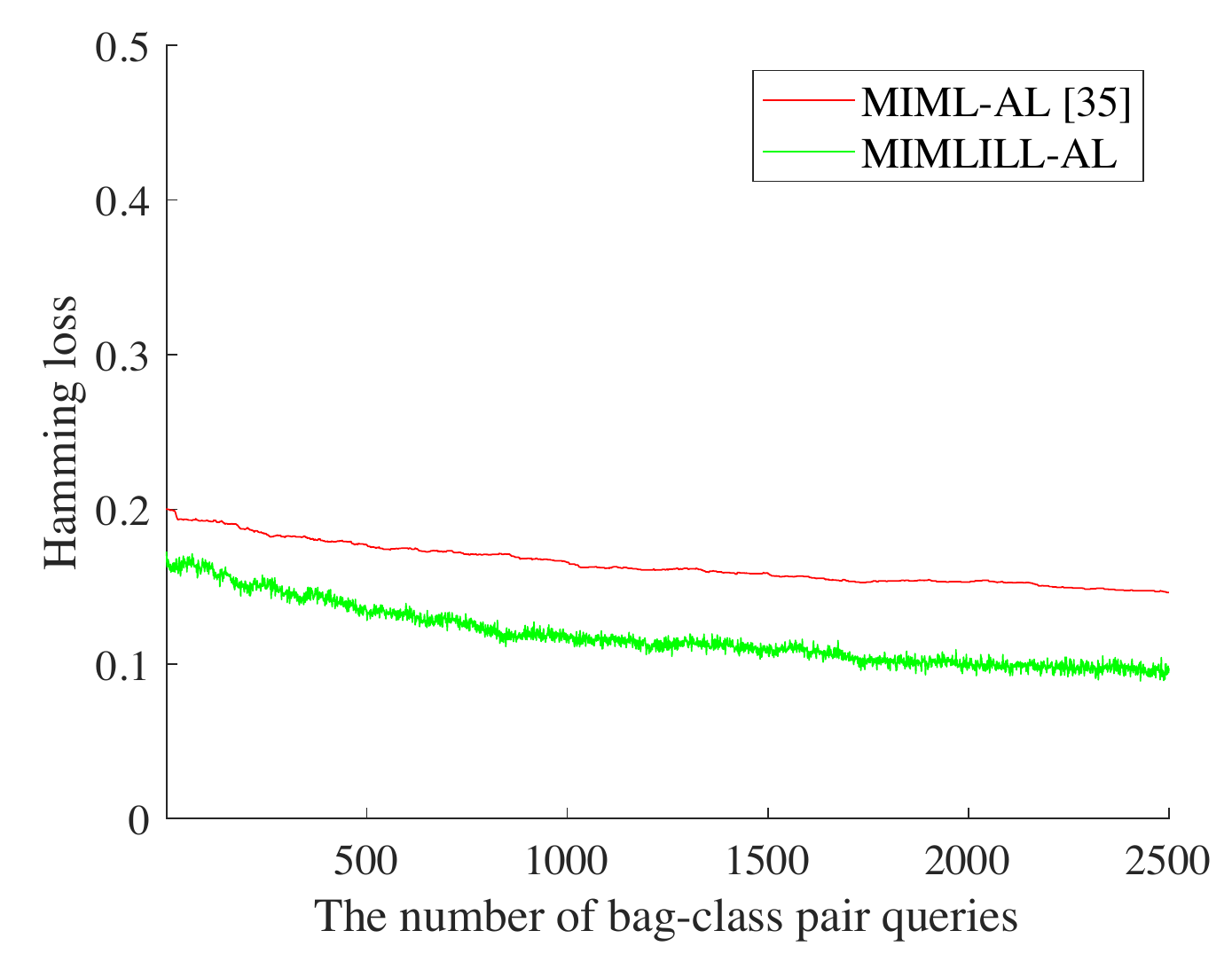}}
    \caption{Carroll}
\end{subfigure}
 \begin{subfigure}{0.33\linewidth}
     \resizebox{\linewidth}{!}{\includegraphics[]{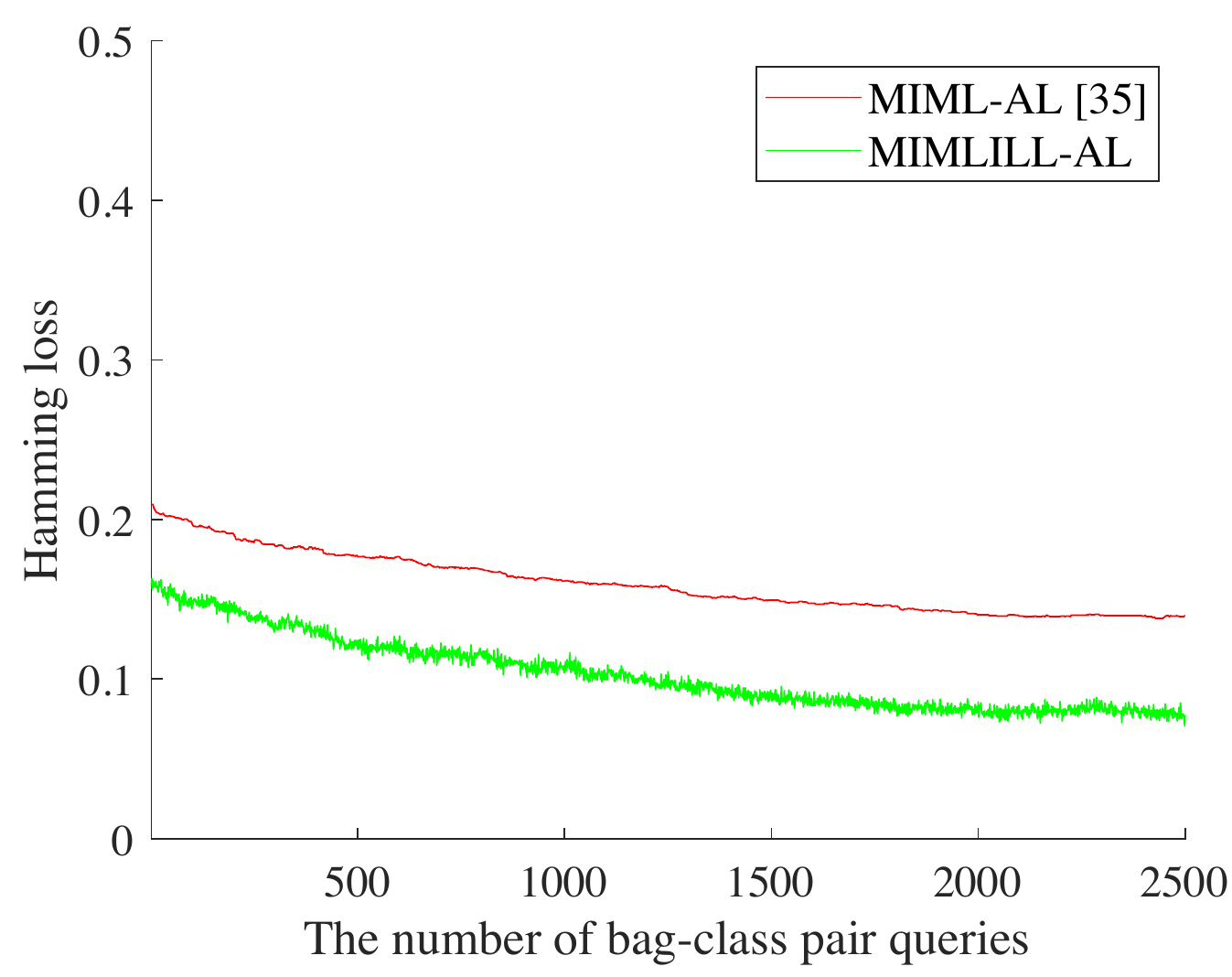}}
    \caption{Frost}
 \end{subfigure}
  \begin{subfigure}{0.33\linewidth}
     \resizebox{\linewidth}{!}{\includegraphics[]{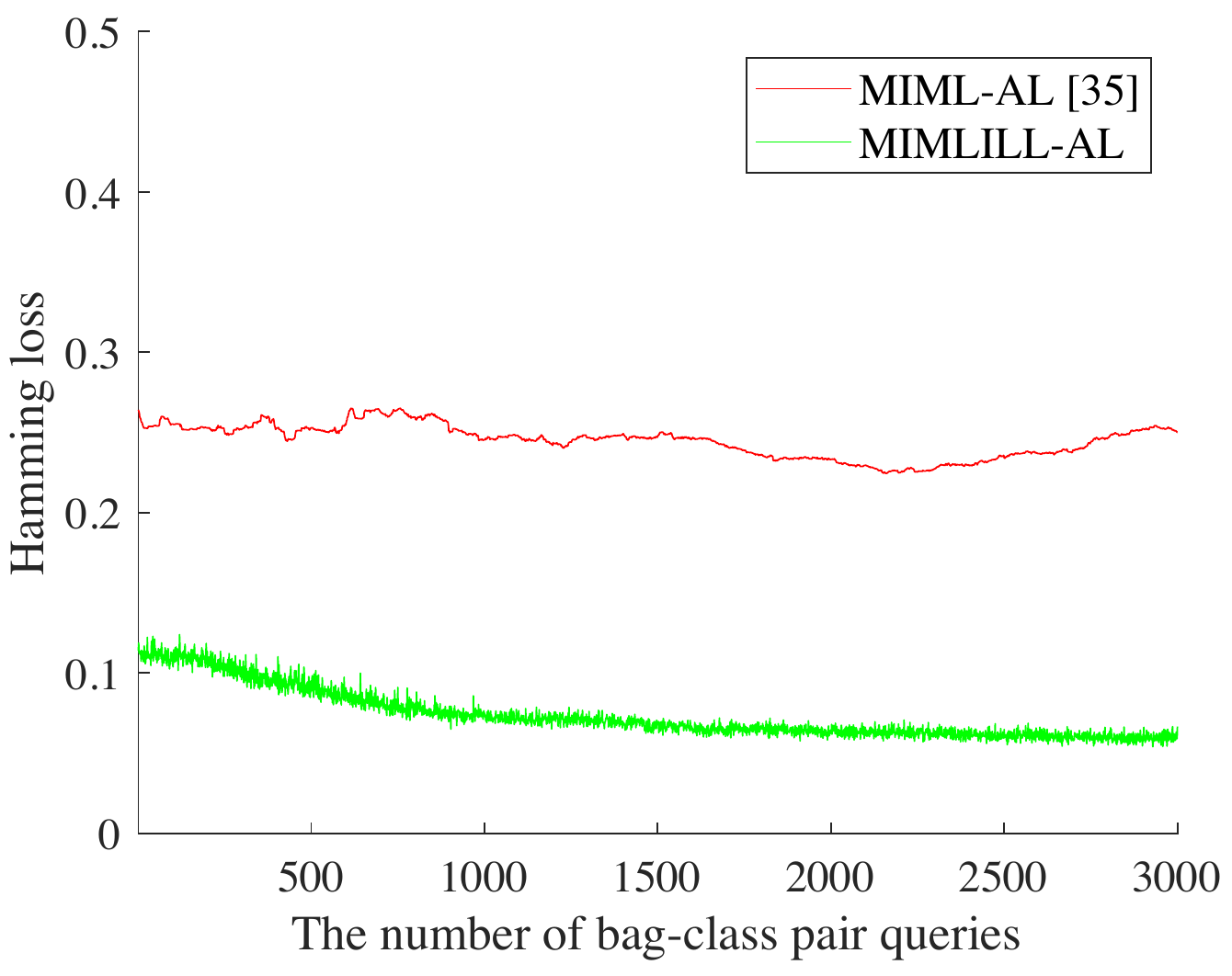}}
    \caption{HJA}
 \end{subfigure}
\caption{ Hamming loss as the function of the number of bag-class pair queries of our method and MIML-AL \cite{huang2017multi} on three datasets: Carroll, Frost, and HJA.} \label{fig:hl_vs_huang}\vspace{-.5em}
\end{figure*}
\begin{figure*}[t!]
\begin{subfigure}{0.33\linewidth}
    \resizebox{\linewidth}{!}{\includegraphics[]{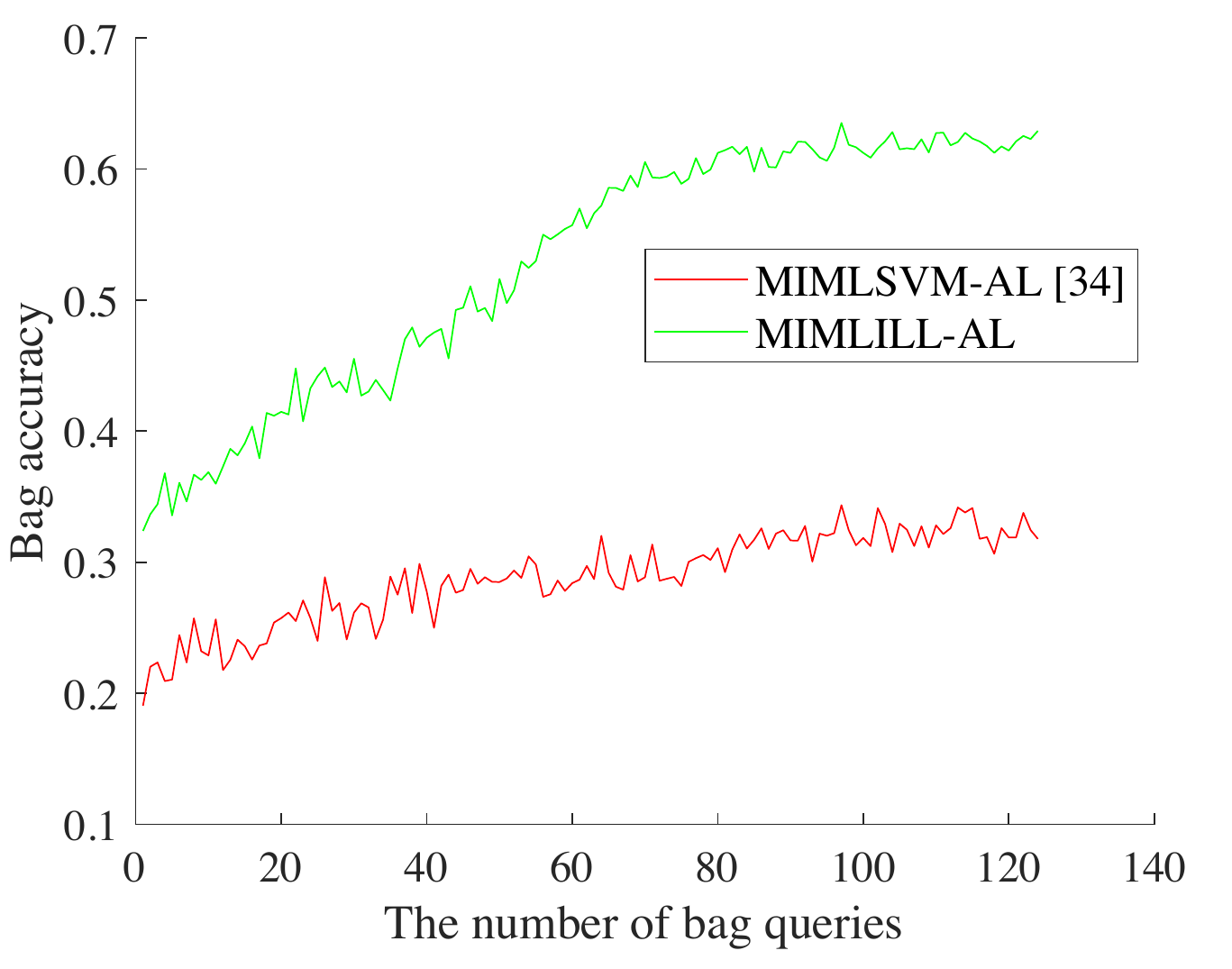}}
    \caption{Carroll}
\end{subfigure}
 \begin{subfigure}{0.33\linewidth}
     \resizebox{\linewidth}{!}{\includegraphics[]{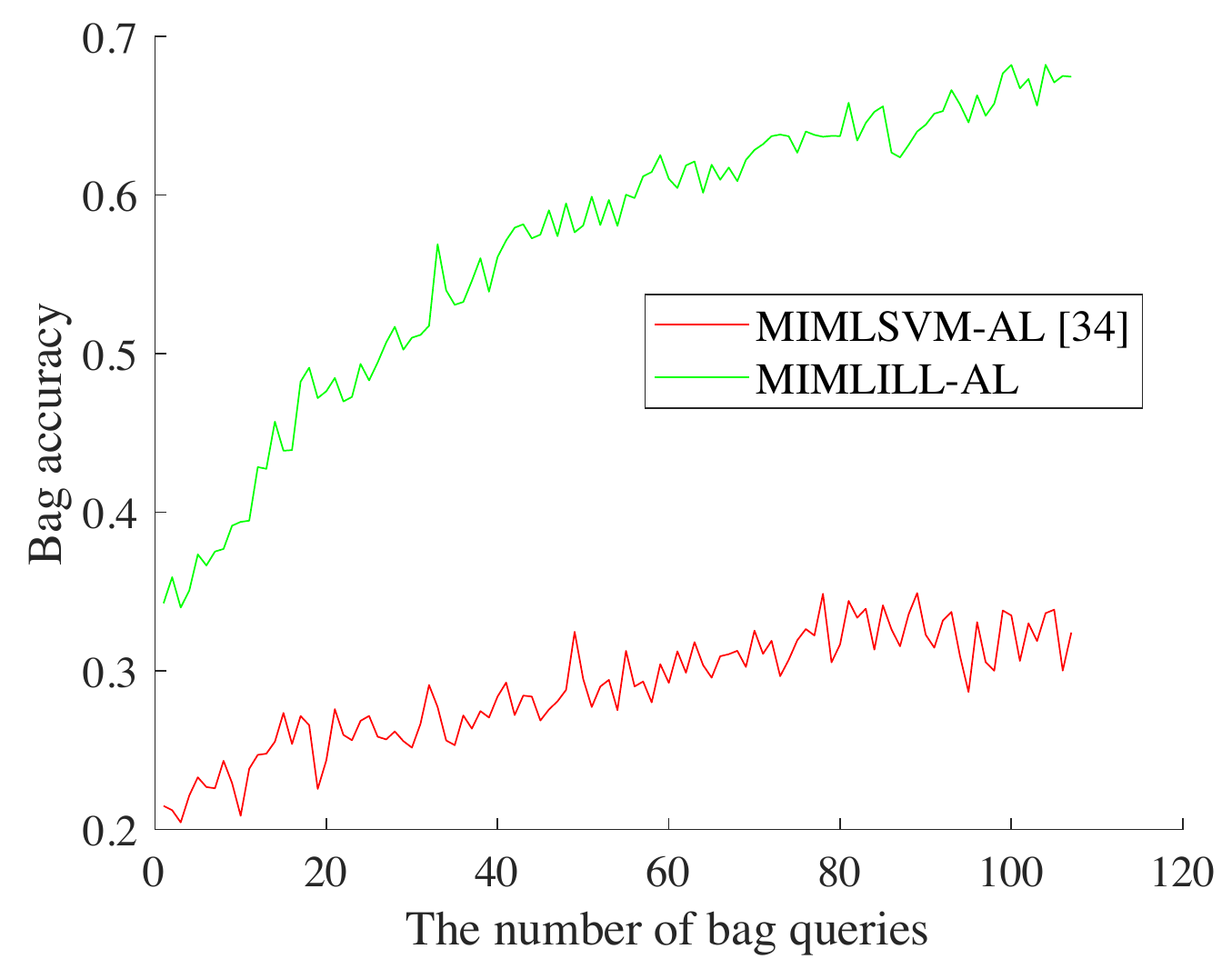}}
    \caption{Frost}
 \end{subfigure}
  \begin{subfigure}{0.33\linewidth}
     \resizebox{\linewidth}{!}{\includegraphics[]{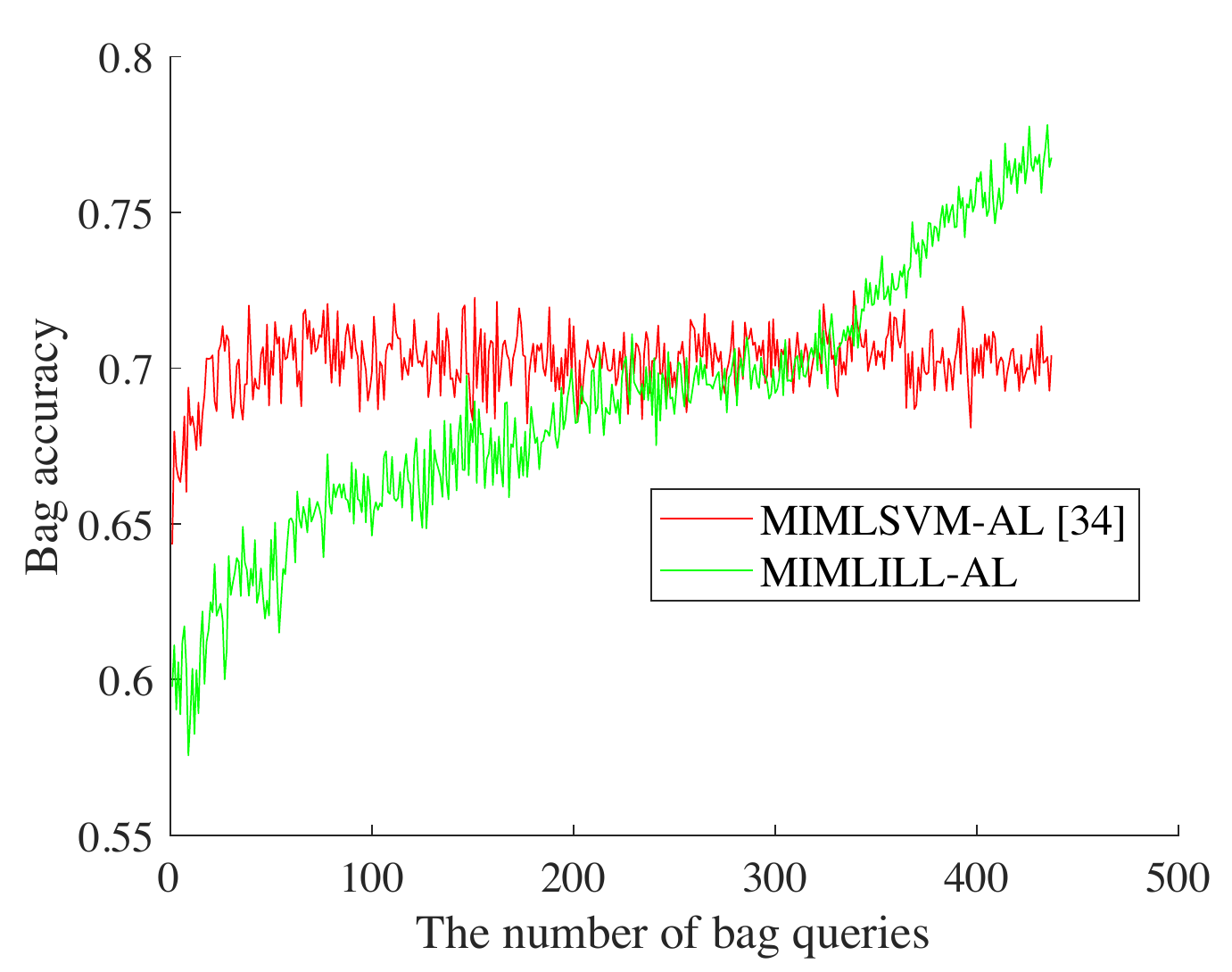}}
    \caption{HJA}
 \end{subfigure}
\caption{ Bag accuracy as the function of the number of bag queries of our method and MIMLSVM-AL \cite{retz2016active} on three datasets: Carroll, Frost, and HJA.} \label{fig:ba_vs_svm}\vspace{-.5em}
\end{figure*}
\begin{figure*}[t!]
\begin{subfigure}{0.33\linewidth}
    \resizebox{\linewidth}{!}{\includegraphics[]{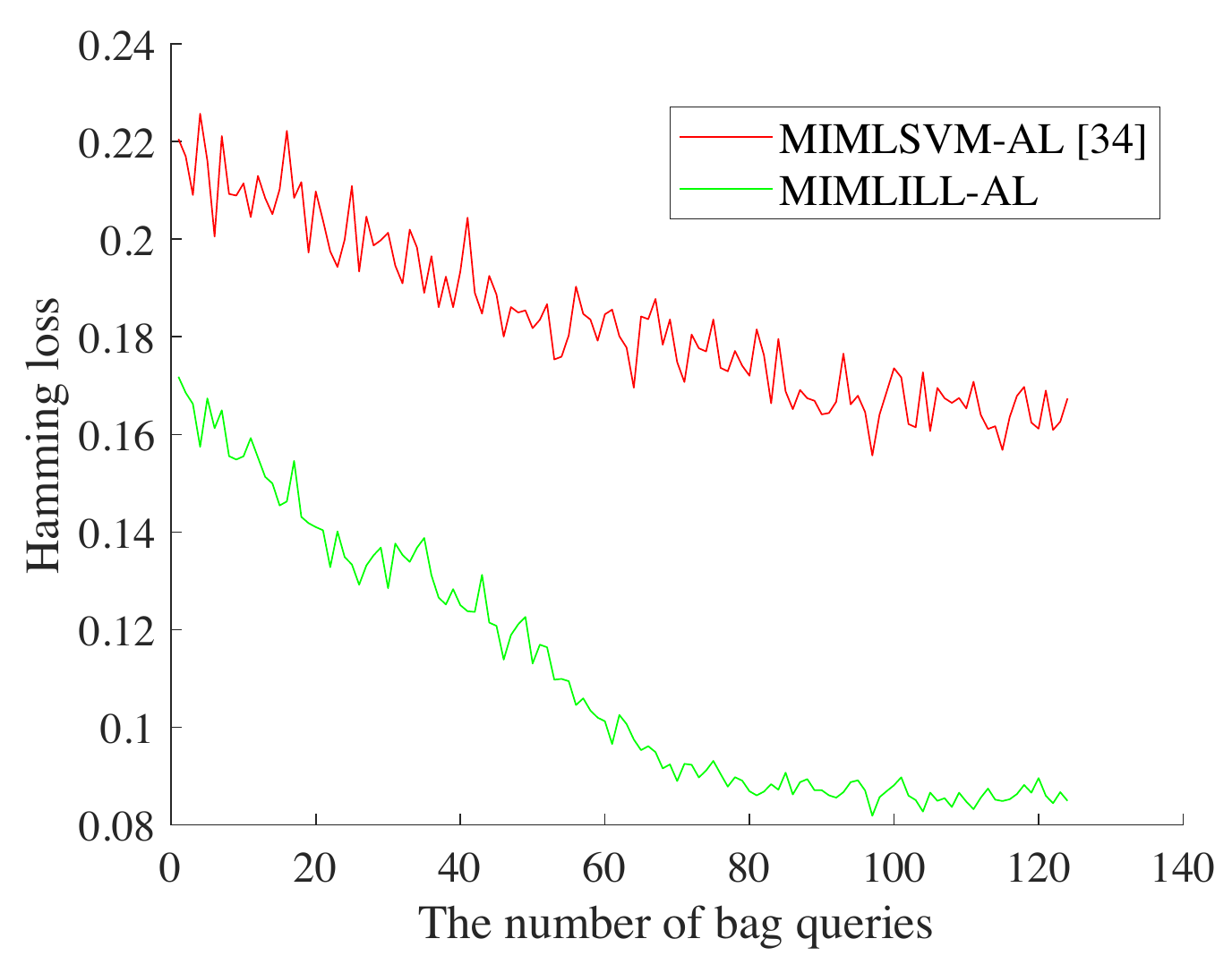}}
    \caption{Carroll}
\end{subfigure}
 \begin{subfigure}{0.33\linewidth}
     \resizebox{\linewidth}{!}{\includegraphics[]{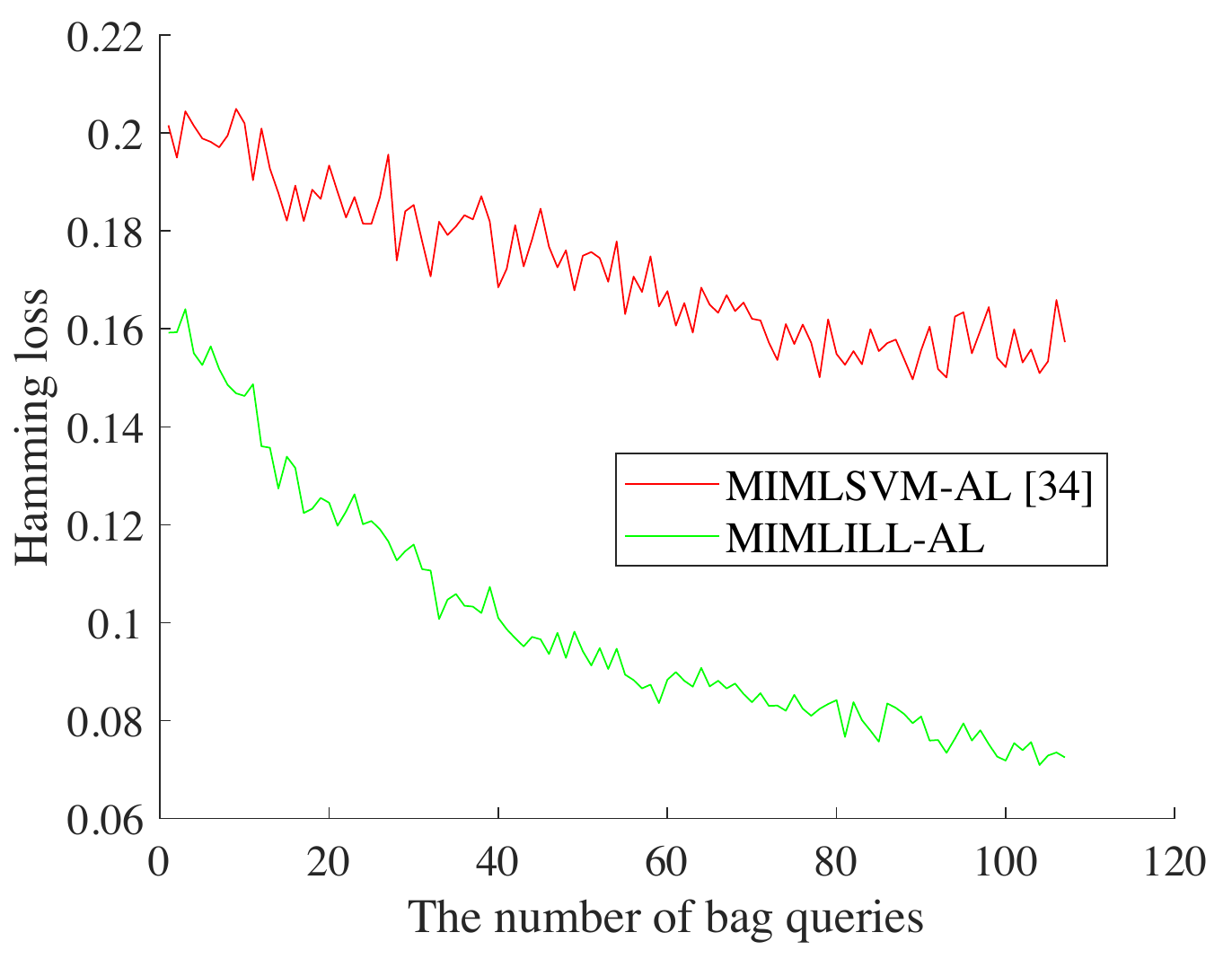}}
    \caption{Frost}
 \end{subfigure}
  \begin{subfigure}{0.33\linewidth}
     \resizebox{\linewidth}{!}{\includegraphics[]{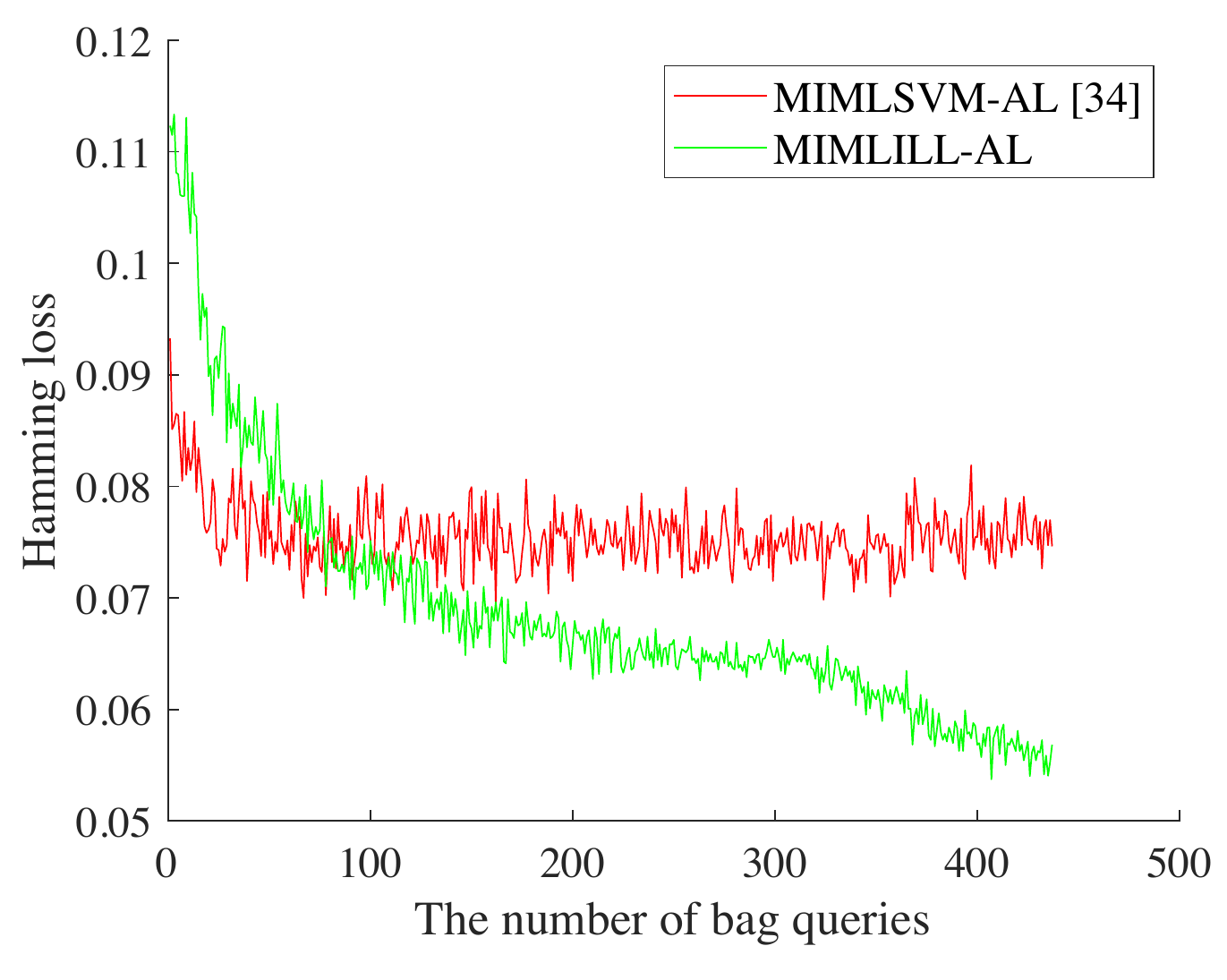}}
    \caption{HJA}
 \end{subfigure}
\caption{Hamming loss as the function of the number of bag queries of our method and MIMLSVM-AL \cite{retz2016active} on three datasets: Carroll, Frost, and HJA.} \label{fig:hl_vs_svm}
\end{figure*} 
\hspace{-2em}\textbf{Result analysis:} The performance of our framework on different selection criteria is presented in Fig.~\ref{fig:bag_accuracy_different_criterion}-Fig.~\ref{fig:oe_different_criterion}. In these figures, the x-axis presents the number of bag-class pair label queries, the y-axis shows the bag accuracy, average precision, Hamming loss, and one error of our method on five different selection criteria. From the curves of five different criteria, we can see that the performance of our two proposed bag-class pair selection criteria based on EGL and uncertainty sampling surpasses the state-of-the-art two selection criteria: bag selection \cite{retz2016active} and bag-then-label \cite{huang2017multi} and the baseline random selection. Our two approaches reach to the performance, which is obtained when all training data is available, very earlier then the rest  methods. Note, in the entire bag selection, after each query, all classes of the selected bag are labeled. To reflect the increase in labeling cost that is proportional to the number of target classes, we plot the performance of each method against the number of bag-class queries rather than queries. When an entire bag is queried, the number of associated individual bag-class queries is $C/k$, where $k$ is the number of classes in a bag that takes the equivalent labeling cost with a single class in the bag. In our illustration, $k=1$, this means that the labeling cost for a single class in a bag equals to the labeling cost for a bag-class pair. We can select $k\geq 1$ when the number of classes is small. Hence, it should be noted that the performance curves in Figs.~\ref{fig:bag_accuracy_different_criterion}-\ref{fig:oe_different_criterion} for bag query methods appear to be sampled less densely than the methods that rely on bag-class queries. \\
\subsection{Comparison with alternative methods}
We compare our model (MIMLILL-AL) with the state-of-the-art algorithms for MIML active learning: MIML-AL \cite{huang2013active} and MIMLSVM-AL \cite{retz2016active}.
On each dataset and on each cross validation, a small part of training data is available, the rest is considered as unlabeled data for active learning. Each query request data to label from the unlabeled data.\\
\textbf{Algorithms:} The method in \cite{retz2016active} is specifically designed based on MIMLSVM. It firstly degenerates bags to single-instance representation and then directly employ traditional active learning method for label querying (select a bag to label all classes every query), which does not truly exploit the characteristics of MIML tasks. 
The authors in \cite{huang2017multi} propose an approach for active learning in MIML setting based on the work in \cite{huang2013active}.  For each query, the bag is selected first based on the uncertainty (the gap between the predicted number of positive labels of the bag and the average number of positive labels of the training data) and diversity of the bag (how many labels of the bag was queried before). Then a class is pointed out to be labeled for the selected bag based on the distance from the label to the thresholding dummy label. In their methods, not only one selected label is queried, but also the key instance which is most relevant to queried label is asked.\\
\noindent\textbf{Evaluation metrics:}
 We report the results based on metrics used for MIML learning evaluation in \cite{tsoumakas2009mining} including: bag accuracy and Hamming loss as the function of the number of queries.\\ 
\textbf{Result analysis:} The performance of our approach and the MIML-AL \cite{huang2017multi} method that is based on bag-then-label selection criterion is shown in Fig.~\ref{fig:ba_vs_huang} and Fig.~\ref{fig:hl_vs_huang}. The x-axis is the number of bag-class pair label queries and the y-axis is bag accuracy or Hamming loss. From these figures, our framework appears superior to MIML-AL in both initialized performance and during querying process. This may be due to the difference of the base classifiers used in our framework and MIML-AL. In MIML-AL, the base classifier is based on fastMIML \cite{huang2018fast}, which constructs a low-dimensional
subspace shared by all labels, and then trains label specific linear models to optimize approximated ranking loss via stochastic gradient descent. Note that the performance of MIML-AL in terms of Hamming loss shown in Fig.~\ref{fig:hl_vs_huang} is comparable to the one reported in MIML-AL paper \cite{huang2017multi} in Carroll and Frost datasets. 
The performance of our approach and the MIMLSVM-AL \cite{retz2016active} method that is based on bag only selection criterion is shown in Fig.~\ref{fig:ba_vs_svm} and Fig.~\ref{fig:hl_vs_svm}. The x-axis is the number of bag queries and the y-axis is bag accuracy or Hamming loss. From these figures, the performance of our framework surpasses the performance of MIMLSVM-AL in Carroll and Frost datasets in terms of bag accuracy and Hamming loss. In case of HJA dataset, the performance of MIMLSVM-AL in terms of Hamming loss and bag accuracy is greater than ours in earlier queries. When the number of queries increases, the performance of our approach reaches and surpasses the performance of MIMLSVM-AL. We suspect that the size of HJA dataset is big enough and the number of classes of HJA dataset is small enough in comparison to Carroll and Frost datasets such that SVM can obtain good classification performance during the initialization step. After that, the performance of MIMLSVM-AL increases gradually when the number of queries increases. The rate of performance increase for MIMLSVM-AL is slower than the rate for our approach. Therefore, our approach achieves better performance in terms of Hamming loss and bag accuracy than MIMLSVM-AL after some queries.
\section{Conclusion}
In this paper, we developed a comprehensive framework for active learning under the MIML-ILL setting. We considered the MIML-ILL model for the classifier use in this paper. To alleviate the computational complexity associated with model update after each query, we developed an online version for maximizing the marginal log-likelihood of the MIML-ILL model. For the query stage, we proposed a novel approach for selecting a bag-class pair by extending EGL and uncertainty sampling to the MIML setting. The experimental evaluation demonstrated the effectiveness and efficiency of the proposed approach.

\ifCLASSOPTIONcaptionsoff
  \newpage
\fi
\bibliographystyle{IEEEtran}
\bibliography{refs}

\begin{thebibliography}{10}
\providecommand{\url}[1]{#1}
\csname url@samestyle\endcsname
\providecommand{\newblock}{\relax}
\providecommand{\bibinfo}[2]{#2}
\providecommand{\BIBentrySTDinterwordspacing}{\spaceskip=0pt\relax}
\providecommand{\BIBentryALTinterwordstretchfactor}{4}
\providecommand{\BIBentryALTinterwordspacing}{\spaceskip=\fontdimen2\font plus
\BIBentryALTinterwordstretchfactor\fontdimen3\font minus
  \fontdimen4\font\relax}
\providecommand{\BIBforeignlanguage}[2]{{%
\expandafter\ifx\csname l@#1\endcsname\relax
\typeout{** WARNING: IEEEtran.bst: No hyphenation pattern has been}%
\typeout{** loaded for the language `#1'. Using the pattern for}%
\typeout{** the default language instead.}%
\else
\language=\csname l@#1\endcsname
\fi
#2}}
\providecommand{\BIBdecl}{\relax}
\BIBdecl

\bibitem{angluin1988queries}
D.~Angluin, ``Queries and concept learning,'' \emph{Machine learning}, vol.~2,
  no.~4, pp. 319--342, 1988.

\bibitem{cohn1996active}
D.~A. Cohn, Z.~Ghahramani, and M.~I. Jordan, ``Active learning with statistical
  models,'' \emph{Journal of artificial intelligence research}, vol.~4, pp.
  129--145, 1996.

\bibitem{king2004functional}
R.~D. King, K.~E. Whelan, F.~M. Jones, P.~G. Reiser, C.~H. Bryant, S.~H.
  Muggleton, D.~B. Kell, and S.~G. Oliver, ``Functional genomic hypothesis
  generation and experimentation by a robot scientist,'' \emph{Nature}, vol.
  427, no. 6971, pp. 247--252, 2004.

\bibitem{king2009automation}
R.~D. King, J.~Rowland, S.~G. Oliver, M.~Young, W.~Aubrey, E.~Byrne,
  M.~Liakata, M.~Markham, P.~Pir, L.~N. Soldatova \emph{et~al.}, ``The
  automation of science,'' \emph{Science}, vol. 324, no. 5923, pp. 85--89,
  2009.

\bibitem{cohn1994improving}
D.~Cohn, L.~Atlas, and R.~Ladner, ``Improving generalization with active
  learning,'' \emph{Machine learning}, vol.~15, no.~2, pp. 201--221, 1994.

\bibitem{dagan1995committee}
I.~Dagan and S.~P. Engelson, ``Committee-based sampling for training
  probabilistic classifiers,'' in \emph{Machine Learning Proceedings
  1995}.\hskip 1em plus 0.5em minus 0.4em\relax Elsevier, 1995, pp. 150--157.

\bibitem{mitchell1982generalization}
T.~M. Mitchell, ``Generalization as search,'' \emph{Artificial intelligence},
  vol.~18, no.~2, pp. 203--226, 1982.

\bibitem{seung1992query}
H.~S. Seung, M.~Opper, and H.~Sompolinsky, ``Query by committee,'' in
  \emph{Proceedings of the fifth annual workshop on Computational learning
  theory}, 1992, pp. 287--294.

\bibitem{dasgupta2008general}
S.~Dasgupta, D.~J. Hsu, and C.~Monteleoni, ``A general agnostic active learning
  algorithm,'' in \emph{Advances in neural information processing systems},
  2008, pp. 353--360.

\bibitem{krishnamurthy2002algorithms}
V.~Krishnamurthy, ``Algorithms for optimal scheduling and management of hidden
  markov model sensors,'' \emph{IEEE Transactions on Signal Processing},
  vol.~50, no.~6, pp. 1382--1397, 2002.

\bibitem{yu2005svm}
H.~Yu, ``Svm selective sampling for ranking with application to data
  retrieval,'' in \emph{Proceedings of the eleventh ACM SIGKDD international
  conference on Knowledge discovery in data mining}, 2005, pp. 354--363.

\bibitem{fujii1998selective}
A.~Fujii, T.~Tokunaga, K.~Inui, and H.~Tanaka, ``Selective sampling for
  example-based word sense disambiguation,'' \emph{Computational Linguistics},
  vol.~24, no.~4, pp. 573--597, 1998.

\bibitem{thompson1999active}
C.~A. Thompson, M.~E. Califf, and R.~J. Mooney, ``Active learning for natural
  language parsing and information extraction,'' in \emph{ICML}.\hskip 1em plus
  0.5em minus 0.4em\relax Citeseer, 1999, pp. 406--414.

\bibitem{moskovitch2007improving}
R.~Moskovitch, N.~Nissim, D.~Stopel, C.~Feher, R.~Englert, and Y.~Elovici,
  ``Improving the detection of unknown computer worms activity using active
  learning,'' in \emph{Annual Conference on Artificial Intelligence}.\hskip 1em
  plus 0.5em minus 0.4em\relax Springer, 2007, pp. 489--493.

\bibitem{lewis1994sequential}
D.~D. Lewis and W.~A. Gale, ``A sequential algorithm for training text
  classifiers,'' in \emph{SIGIR’94}.\hskip 1em plus 0.5em minus 0.4em\relax
  Springer, 1994, pp. 3--12.

\bibitem{mccallumzy1998employing}
A.~K. McCallumzy and K.~Nigamy, ``Employing em and pool-based active learning
  for text classification,'' in \emph{Proc. International Conference on Machine
  Learning (ICML)}.\hskip 1em plus 0.5em minus 0.4em\relax Citeseer, 1998, pp.
  359--367.

\bibitem{hoi2006large}
S.~C. Hoi, R.~Jin, and M.~R. Lyu, ``Large-scale text categorization by batch
  mode active learning,'' in \emph{Proceedings of the 15th international
  conference on World Wide Web}, 2006, pp. 633--642.

\bibitem{tong2001support}
S.~Tong and E.~Chang, ``Support vector machine active learning for image
  retrieval,'' in \emph{Proceedings of the ninth ACM international conference
  on Multimedia}, 2001, pp. 107--118.

\bibitem{zhang2002active}
C.~Zhang and T.~Chen, ``An active learning framework for content-based
  information retrieval,'' \emph{IEEE transactions on multimedia}, vol.~4,
  no.~2, pp. 260--268, 2002.

\bibitem{yang2003automatically}
J.~Yang \emph{et~al.}, ``Automatically labeling video data using multi-class
  active learning,'' in \emph{Proceedings Ninth IEEE international conference
  on computer vision}.\hskip 1em plus 0.5em minus 0.4em\relax IEEE, 2003, pp.
  516--523.

\bibitem{hauptmann2006extreme}
A.~G. Hauptmann, W.-H. Lin, R.~Yan, J.~Yang, and M.-Y. Chen, ``Extreme video
  retrieval: joint maximization of human and computer performance,'' in
  \emph{Proceedings of the 14th ACM international conference on Multimedia},
  2006, pp. 385--394.

\bibitem{tur2005combining}
G.~Tur, D.~Hakkani-T{\"u}r, and R.~E. Schapire, ``Combining active and
  semi-supervised learning for spoken language understanding,'' \emph{Speech
  Communication}, vol.~45, no.~2, pp. 171--186, 2005.

\bibitem{liu2004active}
Y.~Liu, ``Active learning with support vector machine applied to gene
  expression data for cancer classification,'' \emph{Journal of chemical
  information and computer sciences}, vol.~44, no.~6, pp. 1936--1941, 2004.

\bibitem{salmani2014multi}
K.~Salmani and M.~Sridharan, ``Multi-instance active learning with online
  labeling for object recognition,'' in \emph{The Twenty-Seventh International
  Flairs Conference}, 2014.

\bibitem{zhang2010interactive}
D.~Zhang, F.~Wang, Z.~Shi, and C.~Zhang, ``Interactive localized content based
  image retrieval with multiple-instance active learning,'' \emph{Pattern
  Recognition}, vol.~43, no.~2, pp. 478--484, 2010.

\bibitem{settles2008multiple}
B.~Settles, M.~Craven, and S.~Ray, ``Multiple-instance active learning,'' in
  \emph{Advances in neural information processing systems}, 2008, pp.
  1289--1296.

\bibitem{li2013active}
X.~Li and Y.~Guo, ``Active learning with multi-label svm classification,'' in
  \emph{Twenty-Third International Joint Conference on Artificial
  Intelligence}, 2013.

\bibitem{tang2011semantic}
J.~Tang, Z.-J. Zha, D.~Tao, and T.-S. Chua, ``Semantic-gap-oriented active
  learning for multilabel image annotation,'' \emph{IEEE Transactions on Image
  Processing}, vol.~21, no.~4, pp. 2354--2360, 2011.

\bibitem{li2004multilabel}
X.~Li, L.~Wang, and E.~Sung, ``Multilabel svm active learning for image
  classification,'' in \emph{2004 International Conference on Image Processing,
  2004. ICIP'04.}, vol.~4.\hskip 1em plus 0.5em minus 0.4em\relax IEEE, 2004,
  pp. 2207--2210.

\bibitem{hung2011multi}
C.-W. Hung and H.-T. Lin, ``Multi-label active learning with auxiliary
  learner,'' in \emph{Asian conference on machine learning}, 2011, pp.
  315--332.

\bibitem{wu2014multi}
J.~Wu, V.~S. Sheng, J.~Zhang, P.~Zhao, and Z.~Cui, ``Multi-label active
  learning for image classification,'' in \emph{2014 IEEE International
  Conference on Image Processing (ICIP)}.\hskip 1em plus 0.5em minus
  0.4em\relax IEEE, 2014, pp. 5227--5231.

\bibitem{qi2008two}
G.-J. Qi, X.-S. Hua, Y.~Rui, J.~Tang, and H.-J. Zhang, ``Two-dimensional active
  learning for image classification,'' in \emph{2008 IEEE Conference on
  Computer Vision and Pattern Recognition}.\hskip 1em plus 0.5em minus
  0.4em\relax IEEE, 2008, pp. 1--8.

\bibitem{huang2013active}
S.-J. Huang and Z.-H. Zhou, ``Active query driven by uncertainty and diversity
  for incremental multi-label learning,'' in \emph{2013 IEEE 13th International
  Conference on Data Mining}.\hskip 1em plus 0.5em minus 0.4em\relax IEEE,
  2013, pp. 1079--1084.

\bibitem{retz2016active}
R.~Retz and F.~Schwenker, ``Active multi-instance multi-label learning,'' in
  \emph{Analysis of Large and Complex Data}.\hskip 1em plus 0.5em minus
  0.4em\relax Springer, 2016, pp. 91--101.

\bibitem{huang2017multi}
S.-J. Huang, N.~Gao, and S.~Chen, ``Multi-instance multi-label active
  learning.'' in \emph{IJCAI}, 2017, pp. 1886--1892.

\bibitem{wu2018predicting}
J.~Wu, W.~Zhu, Y.~Jiang, G.~Sun, and Y.~Gao, ``Predicting protein functions of
  bacteria genomes via multi-instance multi-label active learning,'' in
  \emph{2018 IEEE 3rd International Conference on Integrated Circuits and
  Microsystems (ICICM)}.\hskip 1em plus 0.5em minus 0.4em\relax IEEE, 2018, pp.
  302--307.

\bibitem{briggs2012acoustic}
F.~Briggs, B.~Lakshminarayanan, L.~Neal, X.~Z. Fern, R.~Raich, S.~J. Hadley,
  A.~S. Hadley, and M.~G. Betts, ``Acoustic classification of multiple
  simultaneous bird species: A multi-instance multi-label approach,'' \emph{The
  Journal of the Acoustical Society of America}, pp. 4640--4650, 2012.

\bibitem{nguyen2020incomplete}
T.~Nguyen and R.~Raich, ``Incomplete label multiple instance multiple label
  learning,'' \emph{IEEE Transactions on Pattern Analysis and Machine
  Intelligence}, 2020.

\bibitem{settles2009active}
B.~Settles, ``Active learning literature survey,'' University of
  Wisconsin-Madison Department of Computer Sciences, Tech. Rep., 2009.

\bibitem{shalev2011pegasos}
S.~Shalev-Shwartz, Y.~Singer, N.~Srebro, and A.~Cotter, ``Pegasos: Primal
  estimated sub-gradient solver for svm,'' \emph{Mathematical programming},
  vol. 127, no.~1, pp. 3--30, 2011.

\bibitem{briggs2012rank}
F.~Briggs, X.~Z. Fern, and R.~Raich, ``Rank-loss support instance machines for
  miml instance annotation,'' in \emph{Proceedings of the 18th ACM SIGKDD
  international conference on Knowledge discovery and data mining}.\hskip 1em
  plus 0.5em minus 0.4em\relax ACM, 2012, pp. 534--542.

\bibitem{tsoumakas2009mining}
G.~Tsoumakas, I.~Katakis, and I.~Vlahavas, ``Mining multi-label data,'' in
  \emph{Data Mining and Knowledge Discovery Handbook}, 2009, pp. 667--685.

\bibitem{huang2018fast}
S.-J. Huang, W.~Gao, and Z.-H. Zhou, ``Fast multi-instance multi-label
  learning,'' \emph{IEEE transactions on pattern analysis and machine
  intelligence}, vol.~41, no.~11, pp. 2614--2627, 2018.

\end{thebibliography}

\clearpage
\pagenumbering{arabic}
\newgeometry{onecolumn,textwidth=\textwidth}
\noindent {\Large \bf Supplemental Material- ``Active Learning in Incomplete Label Multiple Instance Multiple Label Learning", Tam Nguyen and Raviv Raich.}
\begin{appendices}
\renewcommand{\thesectiondis}[2]{\Alph{section}.}
\section{Bounding the optimal parameter vector}\label{appendix:bound_norm}
In the following, we derive a bound on the $l_2$-norm of the parameter vector $\vec{w}$. We begin by expressing the regularized negative marginal log-likelihood objective as follows:
\begin{equation}\label{eq:30}
    f(\vec{w}) = f_0(\vec{w}) + \frac{\lambda}{2}\|\vec{w}\|^2
\end{equation}
where $f_0(\vec{w})$ is the negative marginal log-likelihood objective function, $\lambda$ is the quadratic regularization parameter, and $\| \cdot \|$ denotes the $l_2$-norm. Specifically the negative marginal log-likelihood is given by 
\begin{equation}\label{eq:31}
    f_0(\vec{w})= \frac{1}{\sum_b|S_b|}\sum_{b=1}^B \sum_{c\in S_b}f^o_{bc}(\vec{w},Y_{bc})
\end{equation}
where
\begin{equation}\label{eq:32}
\begin{aligned}
    f^o_{bc}(\vec{w},l) =   -I(l=0) \sum_{i=1}^{n_b} \log P(y_{bi} \neq c) 
     - I(l=1) \log  (1 - e^{\sum_{i=1}^{n_b} \log P(y_{bi} \neq c)}).
\end{aligned}
\end{equation}
Let $\vec{w}^*$ the model parameter vector, which minimizes $f(\vec{w})$. Consequently, we have
\begin{equation}\label{eq:33}
\begin{aligned}
    f(\vec{w}^*) \leq f(\vec{w}) \qquad  \forall \vec{w}.
\end{aligned}
\end{equation}
Since (\ref{eq:33}) holds for any $\vec{w}$, replacing $\vec{w}=0$ into (\ref{eq:33}) and replacing $f(\vec{w}^*)$ with the RHS of (\ref{eq:30}) with $\vec{w}^*$ in place of $\vec{w}$  yields
\begin{equation}\label{eq:34}
\begin{aligned}
    f_0(\vec{w}^*) + \lambda \frac{\|\vec{w^*}\|^2}{2}\leq f(\vec{0})=f_0(\vec{0}).
\end{aligned}
\end{equation}
Reorganizing (\ref{eq:34}), we obtain
\begin{equation}\label{eq:35}
\begin{aligned}
\|\vec{w^*}\| \leq \sqrt{\frac{2}{\lambda} (f_0(\vec{0}) - f_0(\vec{\vec{w}^*}))}.
\end{aligned}
\end{equation}
Since $f_0(\vec{w}^*) \ge 0$, we can upper bound the RHS of (\ref{eq:35}) by $\textstyle{\sqrt{\frac{2}{\lambda} f_0(\vec{0}) }}$ and obtain the following bound on $\| \vec{w}^*\|$:
\begin{equation}\label{eq:36}
\begin{aligned}
\|\vec{w^*}\| \leq \sqrt{\frac{2}{\lambda} f_0(\vec{0}) }.
\end{aligned}
\end{equation}
Next, we proceed by bounding $f_0(\vec{0})$ to further simplify the RHS of (\ref{eq:36}).
Substituting
\begin{equation}
    P(y_{bi}\neq c|\vec{w})|_{\vec{w}=0} = \frac{C-1}{C}.
\end{equation}
into (\ref{eq:32}), we obtain
\begin{equation}\label{eq:f0}
    f^o_{bc}(\vec{0},l) = (1-I(l=1))K_b + I(l=1)(- \log(1-e^{-K_b})), 
\end{equation}
where $\textstyle{K_b =n_b \log \frac{C}{C-1}
}$. We can simplify the bound by replacing the indicators in (\ref{eq:f0}) with the max function as follows:
\begin{equation}\label{eq:f0_max}
    f^o_{bc}(\vec{0},l) \leq \max(K_b,-\log(1-e^{-K_b})).
\end{equation}
To bound the first term in the max function of  (\ref{eq:f0_max}), we bound $K_b$ as follows: 
\begin{eqnarray}\label{eq:40}
\nonumber K_b & =&  n_b \log \frac{C}{C-1}\\
\nonumber  & \le & \max_b n_b   \log \frac{C}{C-1} \\
 & \le & \frac{\max_b n_b}{C-1} 
\end{eqnarray}
where the last inequality uses $\log(1+x) \le x$ with $
\textstyle{x=\frac{1}{C-1}}$. To bound the second term in the max function of  (\ref{eq:f0_max}), we start by lower bounding $K_b$ by 
$K_b \ge \textstyle{\log\frac{C}{C-1} }$ and then bound $-\log(1-e^{-K_b})$, which is monotonically decreasing in $K_b$ as follows:
\begin{eqnarray}\label{eq:41}
\nonumber    -\log(1-e^{-K_b}) & \leq & -\log(1-e^{-\log\frac{C}{C-1}}) \\
\nonumber    & = & -\log(1-e^{\log\frac{C-1}{C}}) \\
\nonumber    & = & -\log(1-\frac{C-1}{C}) \\
\nonumber    & = & -\log(\frac{C-(C-1)}{C}) \\    
\nonumber    & = & -\log(\frac{1}{C}) \\    
    & = & \log (C).
\end{eqnarray}
Substituting the bounds on the first and second term within the max of (\ref{eq:f0_max}), respectively, (\ref{eq:40}) and (\ref{eq:41}), back into (\ref{eq:f0_max}), we obtain: 
\begin{equation}\label{eq:42}
    f^o_{bc}(\vec{0},l) \leq \max(\log (C),\max_b (n_b) \frac{1}{C-1}).
\end{equation}
Substituting the bound on $f^o_{bc}(\vec{0},l)$ in (\ref{eq:42}) into (\ref{eq:31}), we obtain
\begin{equation}\label{eq:43}
    f_0(\vec{0}) \leq \max(\log (C),\max_b (n_b) \frac{1}{C-1}).
\end{equation}
Finally, by substituting the bound on $f_0(\vec{0})$ in (\ref{eq:43}) in (\ref{eq:36}), we obtain the following bound on the $l_2$-norm of the optimal parameter vector:
\begin{equation}
    \begin{aligned}
    \|\vec{w^*}\|  \leq \sqrt{\frac{2}{\lambda}\max(\log(C),\max_b (n_b)\frac{1}{C-1})}.
    \end{aligned}
\end{equation}
Let $\textstyle{\tau = \sqrt{\frac{2}{\lambda}\max(\log(C),\max_b (n_b) \frac{1}{C-1})}}$, we have $\|\vec{w^*}\| \leq \tau$ and $\tau$ is the bound on the $l_2$-norm of the parameter vector $\vec{w}$. Note that this bound holds regardless of the value of the data.
\end{appendices}
\end{document}